\documentclass[letterpaper, 10 pt, journal, twoside]{IEEEtran}

\usepackage{times}
\usepackage{graphicx}
\usepackage{subcaption}
\usepackage{float}
\usepackage{multirow}
\usepackage{amsmath}
\usepackage{stfloats}
\usepackage{color}
\usepackage{amsmath}
\usepackage{amssymb}
\usepackage{cite}
\usepackage{stfloats}
\usepackage{hyperref}
\hypersetup{hypertex=true,
            colorlinks=true,
            linkcolor=blue,
            anchorcolor=blue,
            citecolor=blue}
\DeclareMathOperator{\arctantwo}{arctan2}

\usepackage{booktabs}
\usepackage{flushend}

\newif\ifdraft
\drafttrue
\ifdraft
     \newcommand{\wm}[1]{\textcolor{magenta}{{[wwm: #1]}}}
\else
    \newcommand{\wm}[1]{}
\fi

\begin{document}

\title{
Targetless 6DoF Calibration of LiDAR and 2D Scanning Radar Based on  Cylindrical Occupancy
}

\author{Weimin Wang$^{1}$, Yu Du$^{2}$, Ting Yang$^{2}$, Yu Liu$^{1}$%
\thanks{$^{1}$Weimin Wang and Yu Liu are with DUT-RU International School of Information Science~\&~Engineering, Dalian University of Technology, Dalian, China
        {\tt\footnotesize \{wangweimin, liuyu8824\}@dlut.edu.cn}}%
\thanks{$^{2}$Yu Du and Ting Yang are with the School of Software Technology, Dalian University of Technology, Dalian, China
        {\tt\footnotesize \{d01u01, yangting\}@mail.dlut.edu.cn}
        }%
}

\maketitle

\begin{abstract}

Owing to the capability for reliable and all-weather long-range sensing, the fusion of LiDAR and Radar has been widely applied to autonomous vehicles for robust perception. 
In practical operation, well manually calibrated extrinsic parameters, which are crucial for the fusion of multi-modal sensors, may drift due to the vibration.
To address this issue, we present a novel targetless calibration approach, termed LiRaCo, for the extrinsic 6DoF calibration of LiDAR and Radar sensors.
Although both types of sensors can obtain geometric information, bridging the geometric correspondences between multi-modal data without any clues of explicit artificial markers is nontrivial, mainly due to the low vertical resolution of scanning Radar.
To achieve the targetless calibration, LiRaCo leverages a spatial occupancy consistency between LiDAR point clouds and Radar scans in a common cylindrical representation, considering the increasing data sparsity with distance for both sensors.
Specifically, LiRaCo expands the valid Radar scanned pixels into 3D occupancy grids to constrain LiDAR point clouds based on spatial consistency.
Consequently, a cost function involving extrinsic calibration parameters is formulated based on the spatial overlap of 3D grids and LiDAR points. 
Extrinsic parameters are finally estimated by optimizing the cost function.
Comprehensive quantitative and qualitative experiments on two real outdoor datasets with different LiDAR sensors demonstrate the feasibility and accuracy of the proposed method. The source code will be publicly available.
\end{abstract}

\section{INTRODUCTION}
  
In the advancement of autonomous driving systems, the fusion of various sensors is becoming increasingly indispensable.
The combination of LiDAR~(Light Detection and Ranging) and cameras is a typical setup that leverages the strengths of each sensor in terms of resolution, texture, distant sensing and robustness to lighting conditions.  
However, these visual sensors are susceptible to adverse weather conditions. 
Radar~(Radio Detection and Ranging) sensor, which emits a pulse or a Frequency Modulated Continuous Wave~(FMCW) of radio waves to measure distance, is able to penetrate particles~(e.g., fog, rain) owing to its long wavelengths, but the resulting scanned image is typically low-resolution and noisy due to the width of beam and complexity of signal processing~\cite{radar_different_representation}. 
Currently, the fusion of LiDAR and Radar, which can bring more robust perception even in adverse conditions, is attracting more attention~\cite {MVDnet, ST-mvdNet, RaLiBEV, Bi-LRFusion}.

Similar to the fusion of LiDAR and camera, the extrinsic calibration of LiDAR and Radar sensor performs as an essential and critical step for the fusion of the two sensors. 
Although Radar sensor also captures data mainly based on the geometric information of the environment, the low vertical resolution and the noise of Radar data make it difficult to directly align with LiDAR data. 
In order to determine the calibration parameters 
it is straightforward to measure the relative spatial placement of the sensors manually. 
However, this usually introduces a significant error, mainly due to the lack of identification of the sensors' coordinate system.
Recent calibration methods~\cite{6DoF1,tool1,extrinsic_temporal} utilize the combination of  visual and wave-reflective artificial marker
to design common geometric features that can be identified by both sensors. Even though the sensor set has been well calibrated with target-based methods in the factory, the calibration parameters may drift due to the vibrations or temperature changes during real-world operations. 
In addition, the manual calibration process is usually labor-intensive and time-consuming. 
Therefore, it becomes necessary to investigate the targetless calibration method for LiDAR and Radar sensors to enhance the robustness and facilitate the efficiency for the fusion.

\begin{figure}[tbp]
    \centering
    \includegraphics[width=0.99\linewidth]{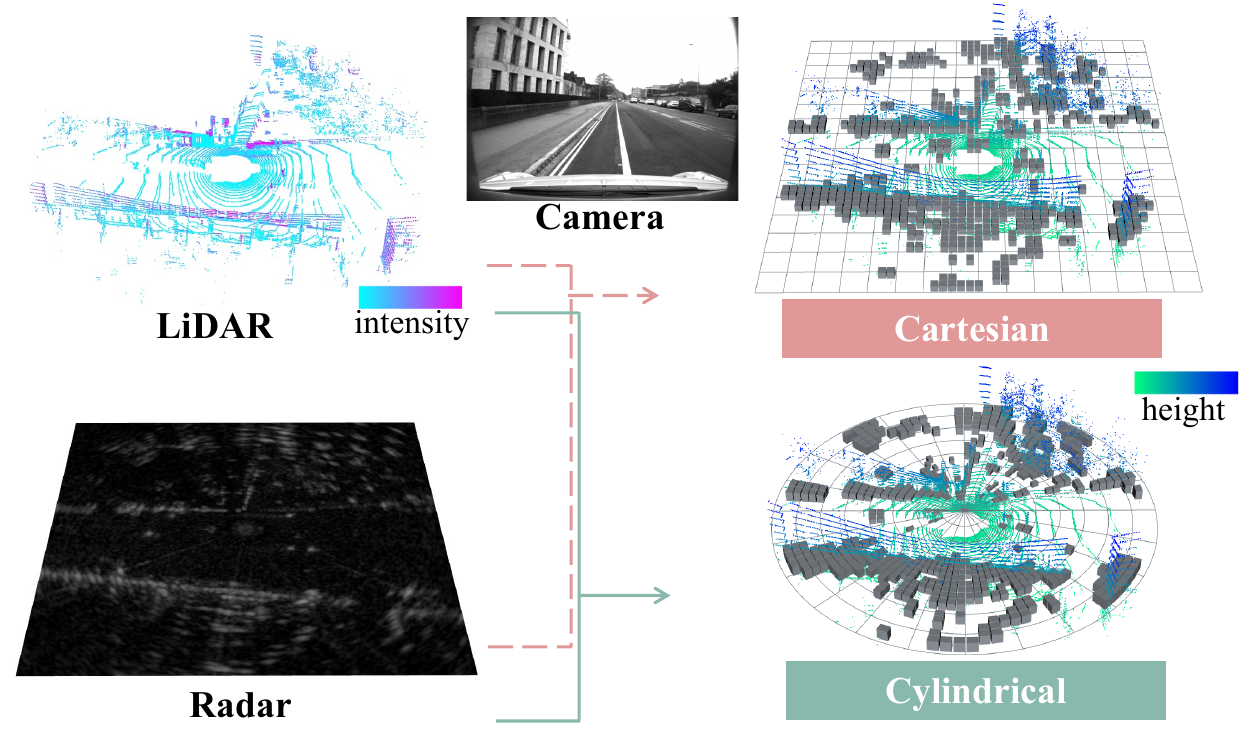}
\caption{
The challenge for LiDAR-Radar calibration and proposed targetless approach. Left: the LiDAR and Radar data of a scene with the RGB image for reference, we can find the difference in physical significance between two sensors in terms of reflective signal and intensity; right: we expand Radar points to 3D areas in cylindrical coordinate system to construct spatial constraints.
}
\label{fig:teaser}
\end{figure}

The key process and challenges for targetless extrinsic calibration involve identifying corresponding features in natural environments. As illustrated in the left part of Fig.~\ref{fig:teaser}, the LiDAR sensor emits multiple narrow laser beams and acquires point-wise 3D coordinates. Conversely, the scanning Radar sensor emits wide FMCW to perceive the 2D distance of the objects within the area covered by the spread lobe. Besides, the reflective intensity of both data is affected by the material but differs in reflective properties. LiDAR is more sensitive to the surface characteristic, whereas Radar primarily responds to the Radar Cross Section~(RCS) of the object. These divergences of physical significance and low vertical resolution of Radar greatly obstruct the establishment of correspondences and make it less constrained to estimate 6 Degrees of Freedom~(DoF) parameters of two sensors.

To address the challenges above, we propose to expand 2D Radar data into 3D areas considering the wave shape of the Radar sensor and constrain LiDAR 3D points by occupancy state, as shown in the right part of Fig.~\ref{fig:teaser}. Specifically, we filter out unoccupied Radar data and expand the scanned pixel into the 3D grid according to the horizontal and vertical resolution. 
To enhance the efficiency and consider the scanning nature of two sensors, the expansion is performed in cylindrical coordinate system,
which is advantageous for handling sparse data in far distances from sensors as  
demonstrated in previous works~\cite{polarnet,nie2023partner}.

Consequently, a cost function that quantifies the spatial occupancy consistency by counting the points in the expanded grids is defined.
Moreover, to tackle the vertical sparsity of the LiDAR data, we introduce the height restrainer that facilitates the transformation of vertically discrete points into a continuous representation.
Finally, we employ the Trust Region Reflective~(TRF) method~\cite{trust-region-methods} to optimize the cost function and estimate the extrinsic parameters. 
The contributions are summarized as follows:
\begin{itemize}
\item We propose \textbf{LiRaCo}, a targetless approach for 6DoF LiDAR-Radar extrinsic calibration. To the best of our knowledge, this is the first work to explore the 6DoF targetless calibration for LiDAR and elevation-less 2D scanning Radar.
\item We formulate a novel cost function based on the expanded cylindrical occupancy grid to effectively bridge the spatial correspondence between LiDAR 3D points and Radar scans and further constrain the 6DoF parameters for optimization.
\item Extensive qualitative and quantitative experiments on two real datasets verify the practicability and superiority of our proposed method over the prior state-of-the-art.
\end{itemize}

\section{Related Work}

\begin{figure*}[h]
\centering
\includegraphics[width=1.0\linewidth]{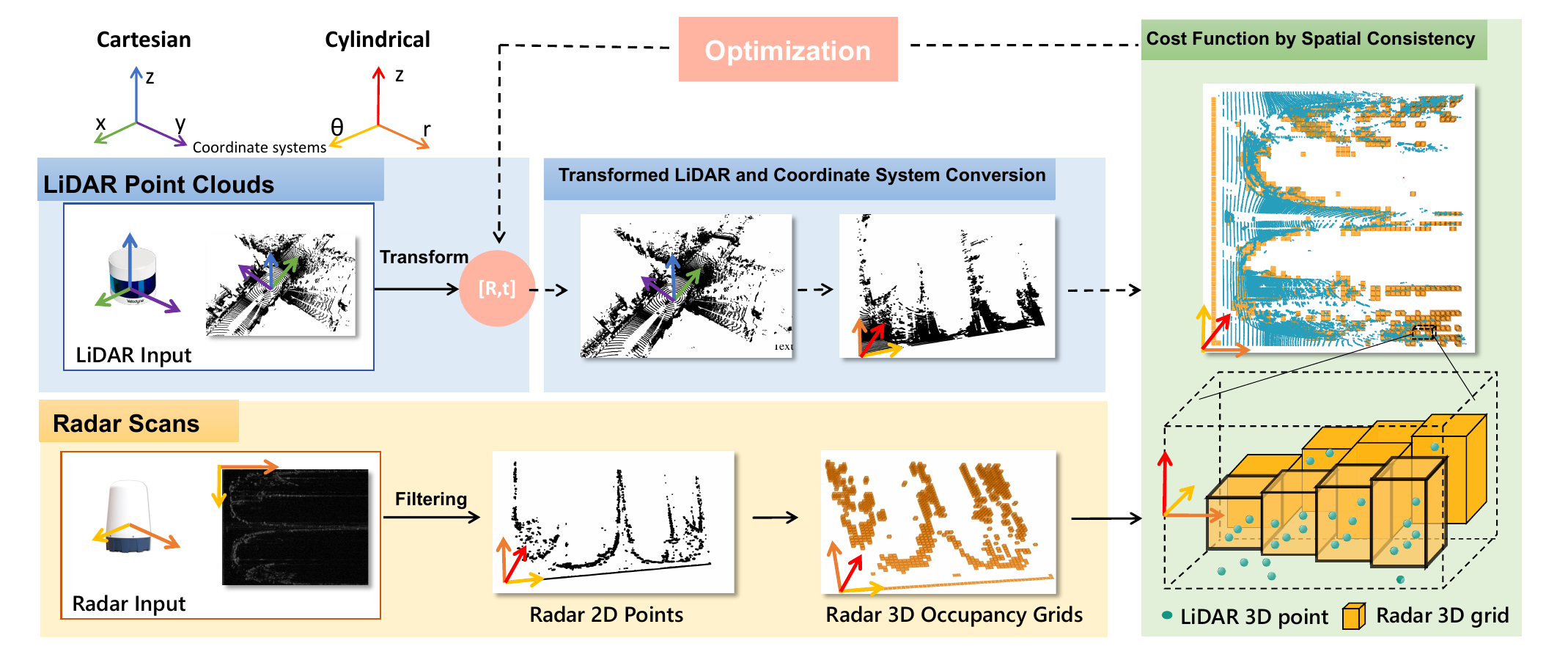}
\caption{The overview of our proposed LiRaCo for targetless LiDAR-Radar extrinsic calibration. Both LiDAR and Radar are represented in cylindrical coordinates. Radar pixels are expanded to 3D occupancy grids to constrain LiDAR points that are transformed with initial or estimated extrinsic parameters. }
\label{fig:architecture}
\end{figure*}

\subsection{LiDAR-Radar Fusion}

Recently, LiDAR fused with 2D scanning Radar sensors is increasingly popular in enhancing perception robustness, owing to the robust sensing capabilities in all weather conditions. Current LiDAR-Radar fusion networks~\cite{MVDnet,ST-mvdNet,ST-MVDNet++,RaLiBEV,Bi-LRFusion,Radar_and_Lidar_Deep_Fusion} primarily use high-quality LiDAR point clouds, incorporating sparse Radar data for augmentation due to its lack of height information.
 MVDNet~\cite{MVDnet} introduced a LiDAR-Radar deep fusion model for vehicle detection, utilizing adaptive feature level attention fusion and demonstrating strong performance under simulated fog.
 RaLiBEV \cite{RaLiBEV} proposed an anchor-free method, improving the generalization and precision of detection. 
 while Bi-LRFusion~\cite{Bi-LRFusion} 
 enriched Radar local features before integrating them.
 
 The impressive performance of these fusion methods lies on reliable extrinsic calibration between two sensors. 
 However, 2D scanning Radar poses challenges for targetless calibration due to its lack of elevation information and distinct sensing modality, necessitating specialized approaches.

\subsection{LiDAR-Radar Calibration}

A straightforward method for extrinsic calibration involves explicitly measuring the sensor's coordinate system~\cite{orr_dataset,radiate_dataset}. However, the ambiguity of the sensor origin constrains the precision of this approach.
Currently, the main extrinsic calibration method is target-based approach~\cite{6DoF1,tool1,extrinsic_temporal,static_auto_calibration_3pairs, Extrinsics_RA_LI_learning,lee2023extrinsic}, similar to the LiDAR-camera calibration~\cite{ILCC,XU2021103776,10341781}. 
These targets meticulously tailored for different sensors' characteristics and application requirements can provide reliable features for detection. 
A notable work on LiDAR-Radar calibration~\cite{6DoF1,6DoF2} designs a calibration target including a triangular trihedral corner reflector and a flat styrofoam triangle board to provide high LiDAR returns and Radar RCS values respectively. 
To jointly calibrate multiple sensors containing LiDAR, Radar and camera, Domhof et al.~\cite{tool1,tool2} utilize a more complex calibration target consisting of a styrofoam board with four circular holes and a metal circular corner reflector; 
however, this method is only evaluated within a 5m distance. 
To address this issue, Agrawal et al.~\cite{static_auto_calibration_3pairs} propose a static roadside infrastructure calibration system based on a simple target-pair which is suitable for short-to-long range calibration.
 
On the other hand, a sensor of an additional modality, such as a camera, can be borrowed as an intermediary to deduce calibration parameters indirectly.
Several calibration works~\cite{2D_radar-camera,3DRadar2ThermalCalib} that focus on Radar and camera are also target-based.
However, for target-based methods, any change in the position or orientation of the sensors typically necessitates re-calibration, thereby increasing the complexity and time cost associated with the calibration target settings.
To eliminate the reliance on specially configured calibration boards, inspired by targetless LiDAR-Camera calibration~\cite{mutual_info,wang2020soic,sem_ins_cali,10356133}, the more generalized calibration method uses notable environmental cues such as trees, roads and cars. 
However, the complexity and variability inherent in real-world settings can affect the accuracy of targetless calibration that relies on natural environmental features, thus this approach often necessitates manual manipulation.
For example, the calibration toolbox Opencalib~\cite{opencalib} provides a targetless Radar2LiDAR tool that relies on manual scene alignment. 
Even though manual annotation is suitable for diverse scenarios, it is labor-intensive and  error-prone when applied to multiple frames with modality gap.

It is still relatively underexplored for the LiDAR-Radar extrinsic calibration without reliance on artificial targets or manual operations.
Heng~\cite{heng2020automatic} pioneered an automatic targetless extrinsic calibration method for multiple Radars and LiDARs, which requires the poses of a moving vehicle. 
This method incorporates the automotive Radar data including range, azimuth, elevation and velocity and defines a cost function by point-to-plane distances between the Radar's 3D points and the constructed LiDAR map, as well as radial velocity. 
Chen et al. recently proposed a motion-based calibration approach ~\cite{iKalibr} for multiple types of sensors including 4D Radar sensors with IMU sensors.
While such Radar sensors provide accurate Doppler velocity and 3D position measurements, making them suitable for ADAS applications,
their limited coverage prevents them from achieving the high resolution and broad area scanning capabilities of the 2D scanning Radar investigated in this work.
However, the 2D scanning Radar faces greater challenges for targetless calibration due to its lack of elevation information. 
Jung et al.~\cite{jung2024imaging} proposed to estimate the  parameters using a generative model to convert 2D scanning Radar bird's-eye view images into LiDAR-style images.
The calibration results of this method are limited to 3DOF only due to the lack of height information for scanning Radar.
To address it, we propose a 6DOF autonomous targetless extrinsic calibration method for LiDAR and $360^{\circ}$ scanning Radar that has no elevation resolution.

\section{METHOD}

\subsection{Overview of LiRaCo} 
Given a set of 3D coordinates of LiDAR point cloud 
$P = \left\{ \textbf{p}_1,...,\textbf{p}_n | \textbf{p}_i=(x_i,y_i,z_i) \in  \mathbb{R}^{3}\right\} $, 
the primary objective of extrinsic calibration is to estimate a rotation 
matrix $\textbf{R}$ and a translation vector $\textbf{t}$ that transform the coordinates from one coordinate system to another, i.e., mapping each point $\textbf{p}^{L}$ of LiDAR's coordinate system to $\textbf{p}^{R}$ of Radar's coordinate system:
\begin{equation}
\textbf{p}^{R} = \textbf{R} \textbf{p}^{L} + \textbf{t}
\end{equation}
where  $\textbf{p} = (x,y,z)$ represents coordinates of a 3D point, $\textbf{R} = (\theta_x, \theta_y, \theta_z ) \in \mathrm{SO}(3)$
and $\textbf{t} = (t_x, t_y, t_z)\in \mathbb{R}^{3}$.
Thus, the essence of the calibration centers on formulating and optimizing a cost function,
which is constrained by spatial correlation and involves the transformation parameters.

As illustrated in Fig.~\ref{fig:architecture}, based on the observation of the spatial correlation between LiDAR points and Radar scans, the pipeline of the proposed LiRaCo mainly consists of three components: transformation and cylindrical conversion of LiDAR points, Radar 3D occupancy grid map generation, and the cost function of spatial correspondence. 
To establish the correspondence between two sensors, we take full advantage of the spatial information by expanding filtered 2D Radar points into 3D occupancy grids to enhance the constraints along the vertical direction.
Moreover, considering the cylindrical scanning nature of two sensors, we propose to build the spatial correspondence in 3D cylindrical coordinates.

\subsection{LiDAR Cylindrical Representation}
\label{chap:3.2}

For the transformed point cloud $\textbf{p}^{R}$ represented in Cartesian coordinate system, we convert it to cylindrical coordinate system with the function denoted as $\Pi(\cdot)$.
\begin{equation}
 \textbf{p}^{R}_{\pi}  = \Pi(\textbf{p}^{R}) 
\end{equation}
where $\textbf{p}^{R}_{\pi}$ = $(r_i,\theta_i,z_i)$.
As a cylindrical coordinate, $z$ remains the value along $Z$-axis, while $r$ indicates the distance on $XY$ plane and $\theta$ denotes the azimuth angle, which are derived by:
\begin{eqnarray}
\left\{
    \begin{aligned}
    r_i &= \sqrt{x_i^2+y_i^2} \\
    \theta_i &= \arctantwo(x_i,y_i) 
    \end{aligned}
    \right.
\end{eqnarray}

\subsection{Radar 3D Occupancy Grid}
\label{chap:3.3}

 The scanned information by Radar is usually stored as a 2D grayscale image in the polar form for a full round sweep over the range, which we denote as $S^{2D}= \left\{ \textbf{s}^{2D}_1,...,\textbf{s}^{2D}_n | \textbf{s}^{2D}_i=(r_i, \theta_i,i_i) \in  \mathbb{R}^{3}\right\} $, where $r, \theta, i$ represent the distance on $XY$ plane, the azimuth angle and the intensity of the reflected signal, respectively. 
 To build the spatial correspondence with 3D LiDAR points, we expand the $Z-$axis dimension and set it as 0 for all $\textbf{s}^{3D}_i=(r_i, \theta_i,0,i_i) $,
 considering the perception principle of the Radar sensor that transmits millimeter waves and receives reflected signals if any reflective object exists in the range of the wave.

\begin{figure}[t]
\centering
\setlength{\belowcaptionskip}{-0.1cm}
 \includegraphics[width=0.96\linewidth]{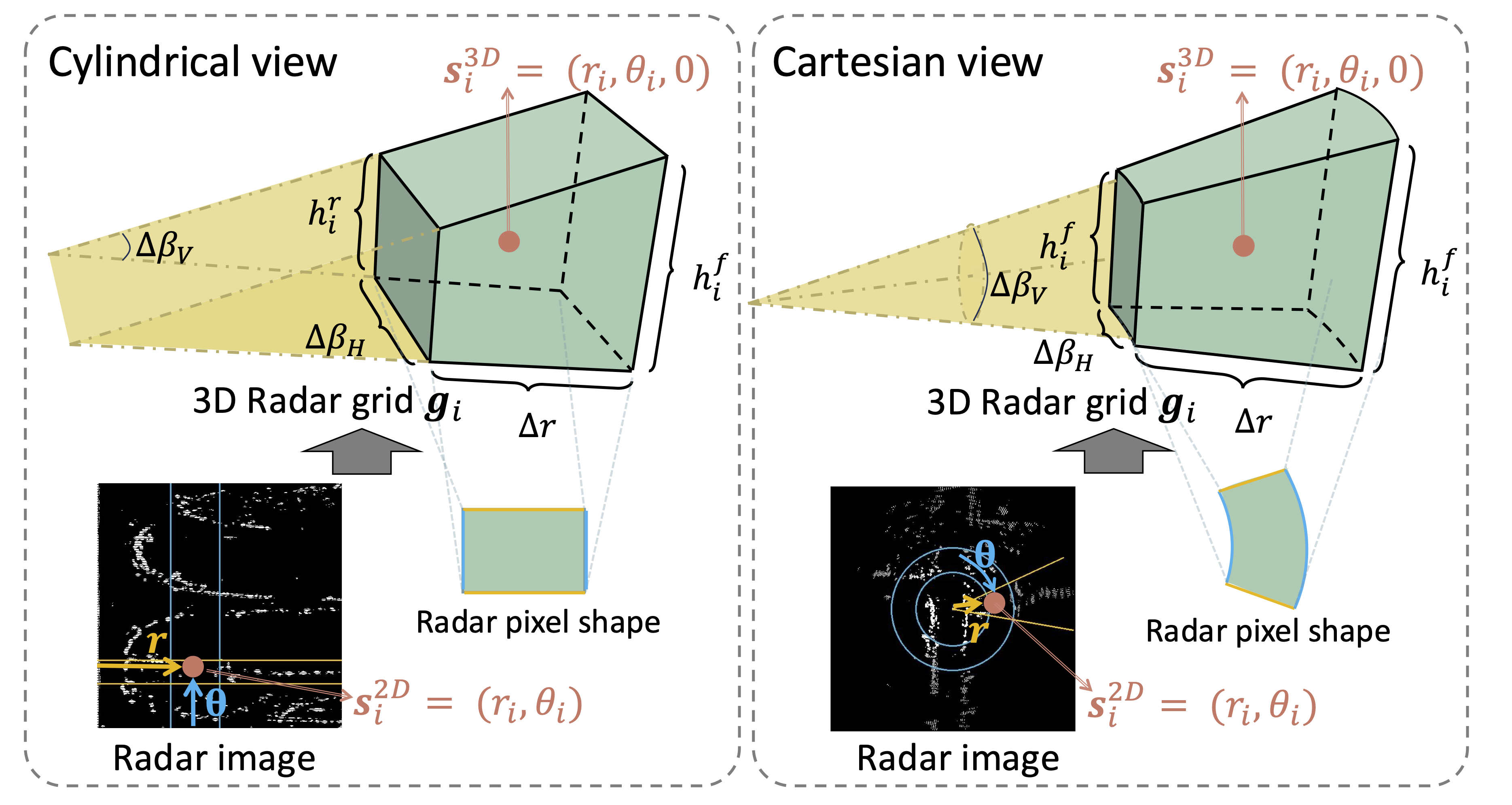}
\caption{
        The illustration of the generated Radar 3D occupancy grid in the cylindrical coordinate system and the reference Cartesian view, considering the form of the Radar wave. The center of the grid is set at $\textbf{s}^{3D}_i=(r_i, \theta_i, 0, i_i)$ and the size of the grid is defined as $(\Delta r,~\Delta \beta_H,~h_i^f,~h_i^r)$ which depend on the range resolution, vertical and horizontal angular resolution and the distance to the coordinate origin.
} 
\label{fig:radar_ocg}
\end{figure}

Moreover, as demonstrated in Fig.~\ref{fig:radar_ocg}, we further define the 3D Radar occupancy grid from each expanded 3D scan $\textbf{s}^{3D}_i$ to increase the spatial overlap with LiDAR for the similar reason. 
Specifically, we denote the 3D occupancy grids by $\mathcal{G} = \{\textbf{g}_1, ... ,\textbf{g}_n  | \textbf{g}_i = (\textbf{s}^{3D}_i, \Delta r, \Delta \beta_H, h_i^f, h_i^r)\}$. Here, the expanded 3D Radar scan $\textbf{s}^{3D}_i$ is the center of the grid, $\Delta r$ relates to the range resolution, $\Delta \beta_H$ is the horizontal angular resolution, and $h_i^f$ and $h_i^r$ are the front and rear heights, which are defined by $r_i$ and the vertical angular resolution $\Delta \beta_V$:
\begin{equation}
\begin{aligned}
h_{i}^f &= 2 \cdot (r_i-\Delta r/2) \cdot \tan(\Delta \beta_V/2)\\
h_{i}^r &= 2 \cdot (r_i+\Delta r/2) \cdot \tan(\Delta \beta_V/2)
\end{aligned}
\label{eq:grid}
\end{equation}

\vspace{-3pt}
\subsection{Cost Function based on Spatial Consistency}
After expanding the Radar scans to 3D occupancy grids, we propose to constrain the target extrinsic parameters with the consideration that the right parameters transform most LiDAR points into occupancy grids. Thus, we define the following cost function $\mathrm{M}(\cdot)$ to quantify the spatial consistency: 
\begin{equation}
    \begin{aligned}
\phi(\textbf{R},\textbf{t}) &=  \sum_{\textbf{p}^{L} \in P^{L}} \mathrm{M} \left( \Pi (\textbf{R} \textbf{p}^{L}+\textbf{t}),\mathcal{G}\right)\\
&=  \sum_{\textbf{p}^{R} \in P^{R}} \mathrm{M} \left( \Pi(\textbf{p}^{R}),\mathcal{G}\right)\\
&=   \sum_{\textbf{p}^{R}_{\pi} \in P^{R}_{\pi}} \mathrm{M}\left(\textbf{p}^{R}_{\pi},\mathcal{G}\right)
    \end{aligned}
\end{equation}
where $\textbf{p}^{L}$ denotes the points in the LiDAR coordinate system,  $\Pi$ is the cylindrical coordinate transformation as described in \S~\ref{chap:3.2} and   $\mathcal{G}$ represents Radar's occupancy grids in \S~\ref{chap:3.3}. Specifically, we consider the following three aspects to compose $\mathrm{M}(\cdot)$.

\textbf{Overlap Indicator.} 
Both LiDAR and Radar sensors yield a valid measurement as a 3D point when the reflectance is caused by interruptions in the laser or mm-wave path. Thus, we define an indicator function $\mathbf {1}$() to quantify this spatial consistency by directly counting the LiDAR points in the Radar's 3D occupancy maps as follows:
\begin{eqnarray}
        \mathbf {1}(\textbf{p}^{R}_{\pi}, \mathcal{G})=\left\{
    \begin{aligned}
    {1},&  & if~\textbf{p}^{R}_{\pi} \in \mathcal{G}  \\
    0,&  & if~\textbf{p}^{R}_{\pi} \notin \mathcal{G} \\
    \end{aligned}
    \right.
\end{eqnarray}

\textbf{Height Restrainer.}
Bridging elevation-sparse LiDAR data with elevation-lacking Radar data poses significant challenges in estimating 6DoF parameters. Although the Radar scans are extended to 3D, it still lacks constraints for parameters related to the vertical dimension, namely, the rotation about $X-$ and $Y-$axis and the translation along $Z-$axis. For example, the cost would not change even the points of a LiDAR's scanline transforms in the range of 3D occupancy map if only the indicator function $\mathrm{1}(\cdot)$ is adopted.
Moreover, when a whole scanline of LiDAR points simultaneously crosses the upper or lower boundaries of the occupancy grids, it results in a discontinuous cost change.

\begin{figure}[!h]
\centering
\setlength{\belowcaptionskip}{-0.1cm}
\includegraphics[width=.4\linewidth]{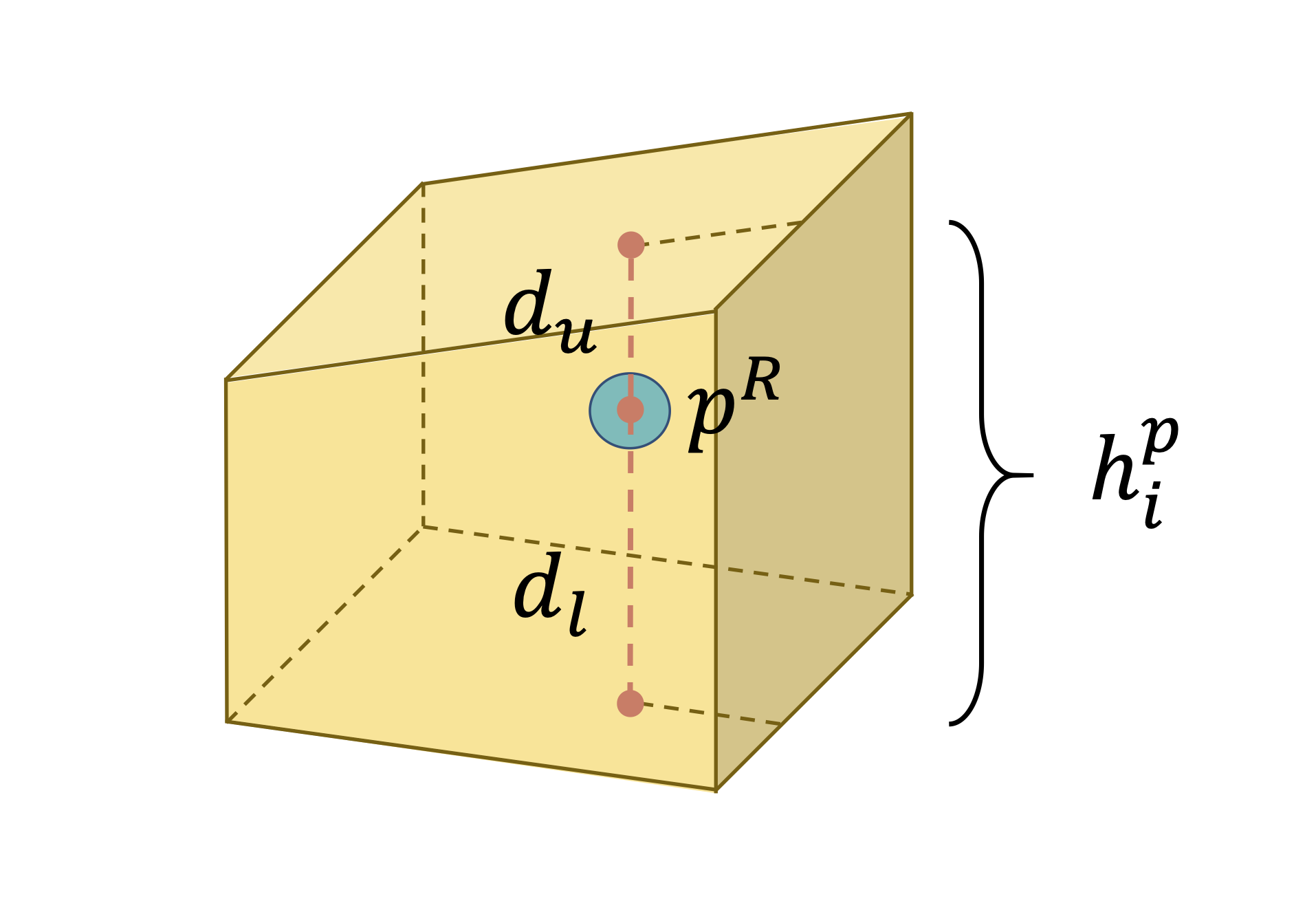}
\caption{The illustration of the height restrainer. LiDAR points are expected to be transformed towards the center of the Radar occupancy grid.
} 
\label{fig:vertical_constr}
\end{figure}

To address the optimization challenge for height-dependent parameters, we define a vertical restrainer $\mathrm{H}(\cdot)$ to enhance the spatial constraints between LiDAR points and Radar's 3D occupancy grid in the vertical direction. The factor could also improve the continuity of the cost function.
As illustrated in Fig.~\ref{fig:vertical_constr}, $\mathrm{H}(\cdot)$  not only considers whether a LiDAR point falls within a 3D occupancy grid, but also takes the distance to the upper and lower face of the occupancy grid as a specific quantitative measurement.
Specifically, $\mathrm{H}(\cdot)$ is defined as: 

\begin{table*}[!h]
\caption{The sensor specification of the ORR and Boreas datasets. }
\label{tab:sensor_specification}
\centering
\begin{tabular}{c|ccccccc}
\hline
Dataset & Sensor & Model & FoV & Resolution & No. & Hz & Range  \\ \hline
\multirow{2}{*}{ORR\cite{orr_dataset}}    & 3D LiDAR   & Velodyne HDL-32E & $360^{\circ}$ $\times$ $41.3^{\circ}$    & 32 channels & 2   &   20   & 100m          \\
& Radar   & Navtech CTS350-X &  $360^{\circ}$  &  $1.8^{\circ}$ beamwidth & 1     &  4    & 163m     \\
                      \hline
\multirow{2}{*}{Boreas\cite{boreas_dataset}}    & 3D LiDAR   & Velodyne Alpha-Prime & $360  ^{\circ}$ $\times$ $40^{\circ}$ & 128 channels &1      &  10    &   180m (5\% reflectivity) 
\\
& Radar   & Navtech CIR304-H  & $360^{\circ}$ &  $1.8^{\circ}$ beamwidth   & 1     &   4   &  200m
\\
\hline
\end{tabular}
\end{table*}

\begin{equation}
\mathrm{H}(\textbf{p}^{R}_{\pi}, \mathcal{G})) = \frac{{(h_{i}^{p})}^2 }
{2 \times (d_u)^2 \times (d_l)^2}
\end{equation}
where $h_{i}^{p}$ represents the height of the occupancy grid $\textbf{g}_i$ where the current LiDAR point $\textbf{p}^{R}_{\pi}$ is located, 
and $d_u =  \frac{h_{i}^{p}}{2} - z$, $d_l = z + \frac{h_{i}^{p}}{2}$ indicate the distance of the LiDAR point to the upper and lower boundaries of the grid.

\vspace{-10pt}

\textbf{Intensity Factor.}
To ensure a high-confidence determination of the grid's occupied state, pixels with low-intensity are considered as background noise and filtered out.
 In addition, we observe that higher intensity of Radar scans indicates wider RCS, meaning more LiDAR points should be generated.
 That is, the scan with high intensity owns a greater probability of the objects' presence, indicating more points should fall into the corresponding extended 3D occupancy grid.
 Thus, we introduce the intensity factor to improve the weights of these key scans, further enhancing the spatial correlation of two modal data.
Specifically, the intensity factor $\mathrm{I}(\textbf{p}^{R}_{\pi}, \mathcal{G})$ is defined as:
\begin{eqnarray}
\mathrm{I}(\textbf{p}^{R}_{\pi}, \mathcal{G})=
    \begin{cases}
    1.5,&  i_{i} > W_{th}  \\
    1,  & W_{th}  \geq i_{i} > V_{th}  \\
    0,  &  otherwise  \\
    \end{cases}
\label{eq:inten}
\end{eqnarray}
where $W_{th}$ denotes the intensity threshold of weight enhanced grids, and $V_{th}$ denotes the intensity threshold of valid occupied grids.

Finally, the cost function is defined by the number of valid LiDAR points within the Radar occupancy map, incorporating spatial and intensity constraints:

\begin{equation}
\mathrm{M} = \mathbf {1}_{} \cdot \mathrm{H}_{} \cdot \mathrm{I}_{}
\end{equation}
\subsection{Optimization}
A higher cost indicates a greater degree of spatial feature overlap between the two sensors, suggesting that the extrinsic parameters are more precisely determined.
Therefore, we can obtain the extrinsic parameters $\hat{\textbf{R}}$ and $\hat{\textbf{t}}$
by minimizing the negative of spatial consistency cost function:

\begin{equation}
\hat{\textbf{R}}, \hat{\textbf{t}} = \arg \min_{\textbf{R},\textbf{t}} (-\phi(\textbf{R},\textbf{t}))
\end{equation} 

We utilize Trust Region Reflective (TRF) algorithm~\cite{trust-region-methods} to optimize the extrinsic parameters due to its advantages in handling bound-constrained problems. TRF starts by calculating the cost with the initial guess. During each iteration, the TRF algorithm constructs a trust region around the current solution and performs a local optimization within this region based on the cost change. This process involves determining the appropriate step size and direction based on local approximations and the dimensions of the trust region.

\section{Experiments}

\subsection{Experimental Setup}
\textbf{Datasets}. To evaluate the proposed extrinsic calibration method, we conduct experiments on two real-world datasets:  Oxford Radar RobotCar~(ORR)~\cite{orr_dataset} and Boreas~\cite{boreas_dataset}. Both datasets are collected using a comprehensive sensor suite, including scanning Radar and 3D LiDAR
as listed in Table \ref{tab:sensor_specification}.
ORR is a pioneering environmental perception dataset that integrates camera, LiDAR, and FMCW scanning Radar,
which consists of 32 sequences in different traffic, weather and lighting conditions.
Boreas is a large multi-modal dataset that focuses on localization, which includes repeated traversals of a route near the University of Toronto.

To mitigate the impact of temporal synchronization, LiRaCo only takes stationary frames of two sensors to estimate the spatial parameters.  Radar frames are considered stationary if the velocity of five consecutive frames is less than $0.05~[m/s]$.
In ORR, we use a total of 101 stationary frames from four clear sequences collected on $17/01/2019$. In Boreas,
each corresponding LiDAR frame is selected with the closest timestamp. A total of 28 stationary pairs are available from the Boreas-Objects-V1 which is a subset of Boreas.
As for the Ground Truth~(GT) extrinsic parameters, ORR dataset performs pose optimization of the co-observations error with the initial seed manually measured as-built positions of the sensors.
The Boreas dataset estimates rotation by averaging the results of several pairs derived using the Fourier Mellin transformation and measures the translation from the CAD model of the roof rack.

\textbf{Data preprocessing}.
The raw Radar scan images are $3768\times400$ with a range-azimuth resolution of $0.0438~[m]\times0.9~[^{\circ}]$ for ORR, and
$3360\times400$ with $0.0596~[m]\times0.9~[^{\circ}]$ for Boreas, respectively.
To conform with the range of LiDAR, we crop the range of Radar image to 100~[m], that is,
the resolutions are changed to $2283\times400$ for ORR and $1680\times400$ for Boreas.

\textbf{Implementation details.}
For the optimization process, we utilize the TRF~\cite{trust-region-methods} implementation in SciPy. The relative step size for the finite difference approximation of the Jacobian is set to $(0.2,0.2,0.1,0.1,0.1,0.1)$ and the tolerance for the termination is configured to 0.001.
We define the lower and upper bounds of variables for optimization as [$-10^\circ, 10^\circ$] for rotation and [-2m, 2m] for translation parameters. For experiments involving the transformed point cloud, these bounds are reduced to half.
The  vertical angular resolution $\Delta \beta_V$ in Eq.~\ref{eq:grid} and horizontal angular resolution $\Delta \beta_H$ of $\mathcal{G}$ are $1.8^{\circ}$ and $0.9^{\circ}$ for the Radar sensor used in this work. Intensity thresholds of $V_{th}$ and $W_{th}$ defined in Eq.~\ref{eq:inten} are 50 and 80 respectively.

\subsection{Baseline Methods}
To evaluate the effectiveness of LiRaCo, we take three methods as baselines for the comparison: Template Matching~(TM)~\cite{opencv_library}, 
Fourier-Mellin Transform~(FMT)~\cite{boreas_dataset}
and Iterative Closest Point~(ICP)~\cite{icp}.
Note that most of these methods can only estimate partial extrinsics, while our proposed LiRaCo is able to estimate all six parameters. 

TM is usually used to search and locate a template pattern of a large area by computing the similarity between the template and the content in the window. In this work, we use TM to 
estimate the optimal offsets between the LiDAR and Radar Bird's Eye View~(BEV) images, namely $t_x$ and $t_y$.
Similarly,
FMT is also a robust method for scaling and 2D rotation transformations, which is unitized to estimate the rotation parameter in Boreas dataset~\cite{boreas_dataset}.
ICP is a widely used method for 3D point cloud registration. To make it applicable to Radar points, we  extend the z-axis with the value set to 0,
which is the same as in \S\ref{chap:3.3}.

\begin{table*}[h]
\caption{Estimated parameters on ORR Dataset. \textbf{$ \theta $}~[$^{\circ}$] and \textbf{$ t$}~[$m$] represent the rotation parameters and translation parameters respectively. ``-'' indicates the incapability of estimating the corresponding parameter. Bold numbers represent the best mean values among the methods.}
\centering
\begin{tabular}{c|cccccc}
\hline
               Methods
               & $\theta _x$ & $\theta _y$ & $ \theta _z$ & $t_x$ &  $t_y$ & $t_z$   \\ \hline
GT             & 0.17   &  0.46  &  0.34  & 0.09   &   0.44 &  0.28  \\ 
\hline
TM~\cite{opencv_library} &  -  & -  & -  & -0.14~(0.69)  & 0.08~(1.01)  & -  \\ 
FMT~\cite{boreas_dataset}             & -   & -  &  38.87~(19.53)  & -   &  -  & -  \\
ICP~\cite{icp}                & 1.94~(0.10)   & -0.75~(0.09)  &  \textbf{0.35}~(0.68)  & -0.37~(0.26)   &  0.09~(0.27)  & -1.00~(0.02)  \\
\hline
LiRaCo~(Ours)           & \textbf{-0.13}~(2.4)   & \textbf{0.02}~(1.70)   &  0.30~(0.34)  & \textbf{0.12}~(0.19)   & \textbf{0.34}~(0.07)   & \textbf{0.10}~(0.23)  \\ 
\hline
\end{tabular}
\label{tab:orr_calibration}
\end{table*}

\begin{table*}[t]
\caption{Calibration Results on Boreas Dataset. 
} 
\centering
\begin{tabular}{c|cccccc}

\hline
               Methods
              & $\theta _x$ & $\theta _y$ & $ \theta _z$ & $t_x$ &  $t_y$ & $t_z$   \\ \hline

GT             & 0.00   &  0.00  &  -2.25  & 0.00   &   0.00 &  0.21  \\ 
\hline
TM~\cite{opencv_library} &  -  & -  & -  & 0.05~(0.31)  & 0.49~(0.61)  & -  \\ 
FMT~\cite{boreas_dataset}            & -   & -  &  32.77~(18.89)  & -   &  -  & -  \\
ICP~\cite{icp}             & -\textbf{0.17}~(0.05)   & -0.22~(0.24)  &  -1.94~(0.64)  & -0.03~(0.20)   &  0.06~(0.16)  & -0.41~(0.01)  \\
\hline
LiRaCo~(Ours)           & -0.21~(0.57)   & \textbf{-0.02}~(0.24)   & \textbf{-2.77}~(0.07)  & \textbf{0.00}~(0.02)   & \textbf{0.02}~(0.03)   & \textbf{0.24}~(0.06)  \\ 
\hline
\end{tabular}
\label{tab:boreas_calibration}
\end{table*}
\begin{figure}[H]
    \centering

    \begin{subfigure}{1\linewidth}
        \setlength{\abovecaptionskip}{0.3cm}
        \setlength{\belowcaptionskip}{0.2cm}
        \centering
         \includegraphics[width=0.302\linewidth]{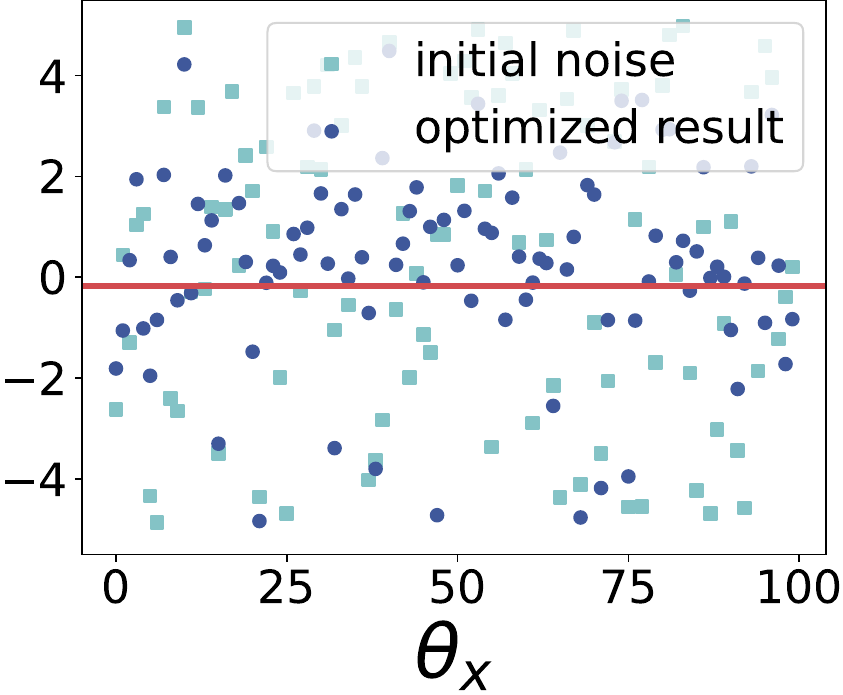}
         \hspace{.7mm}
        \includegraphics[width=0.302\linewidth]{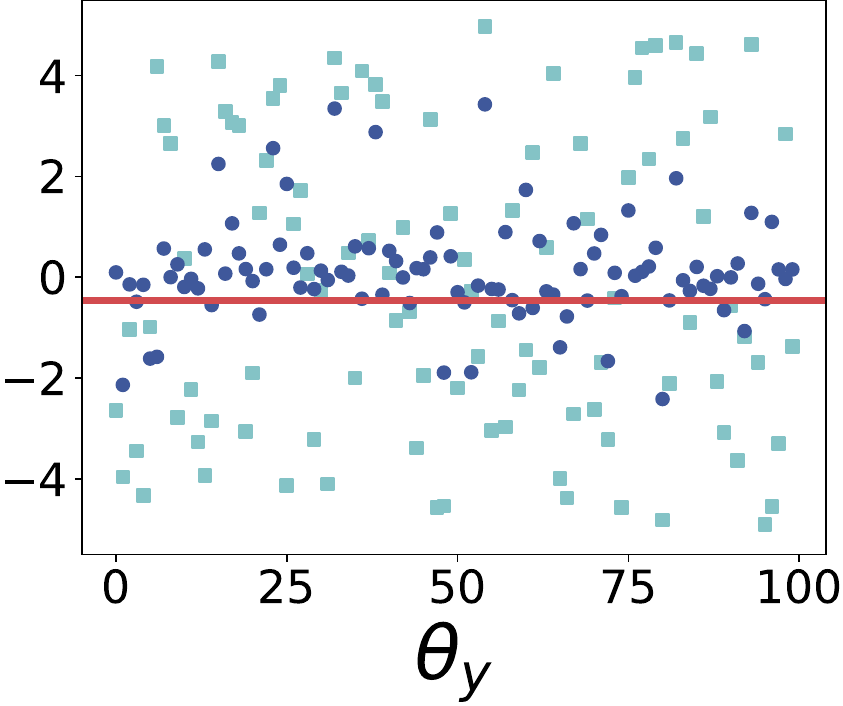}
         \hspace{.7mm}
        \includegraphics[width=0.302\linewidth]{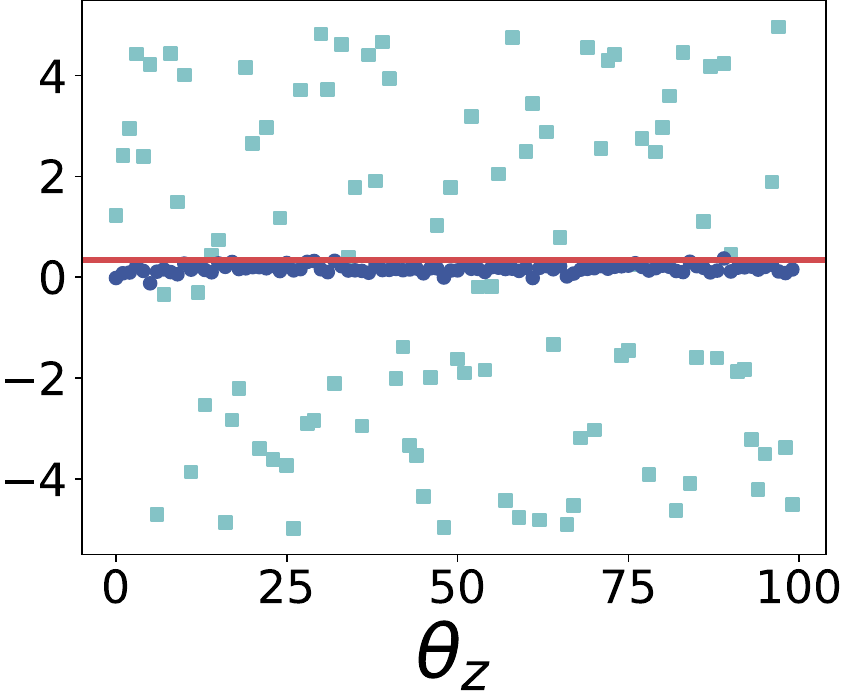}
    \end{subfigure}

    \begin{subfigure}{1\linewidth}
        \centering
        \includegraphics[width=0.32\linewidth]{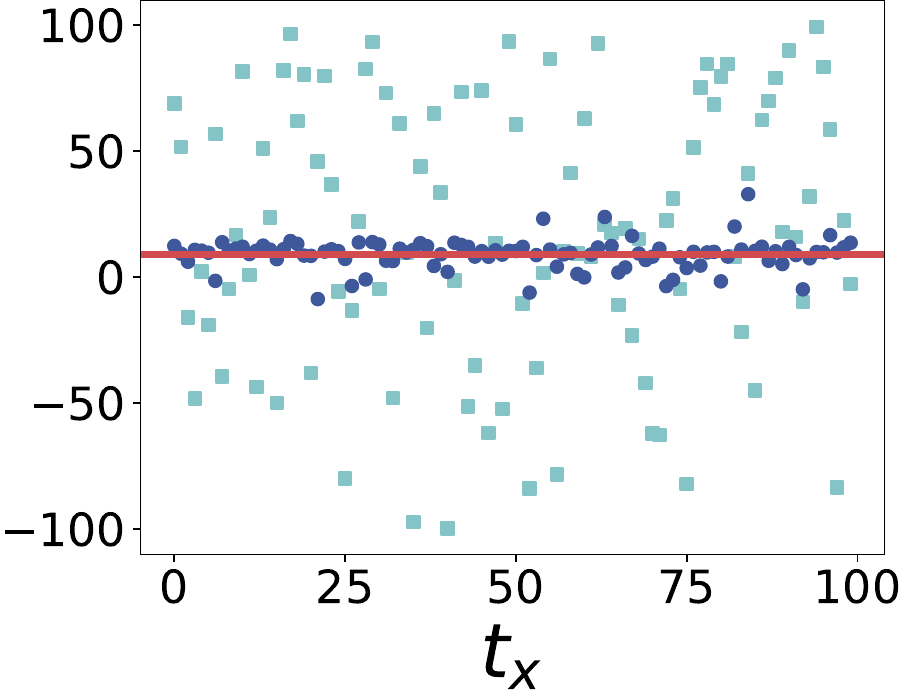}
        \includegraphics[width=0.32\linewidth]{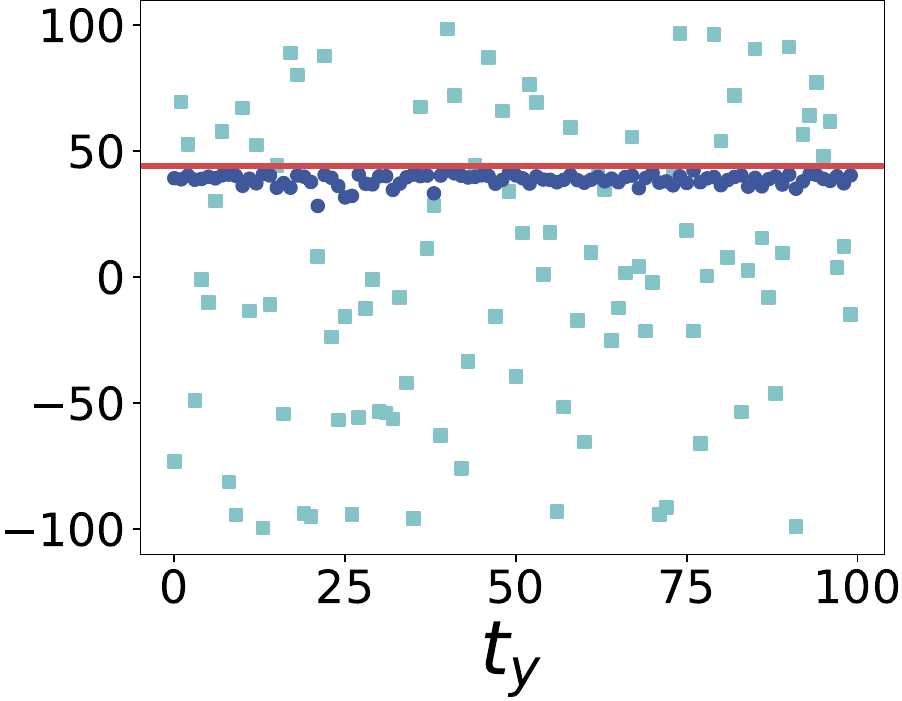}
        \includegraphics[width=0.32\linewidth]{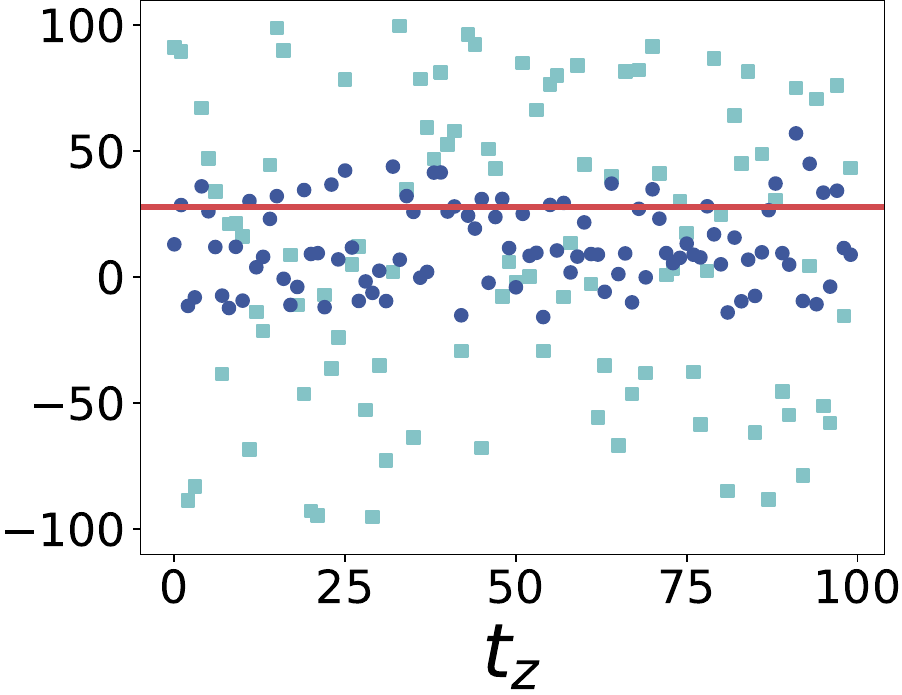}
        \vspace{-.2cm}
        \caption{}
    \end{subfigure}

    \begin{subfigure}{1\linewidth}
        \setlength{\belowcaptionskip}{0.2cm}
        \centering
        \includegraphics[width=0.32\linewidth]{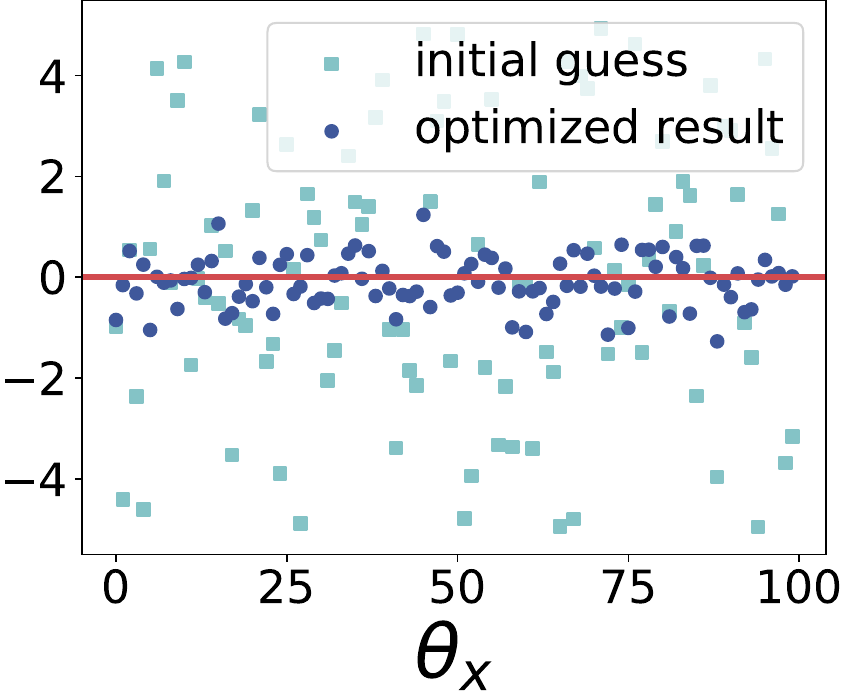}
        \includegraphics[width=0.32\linewidth]{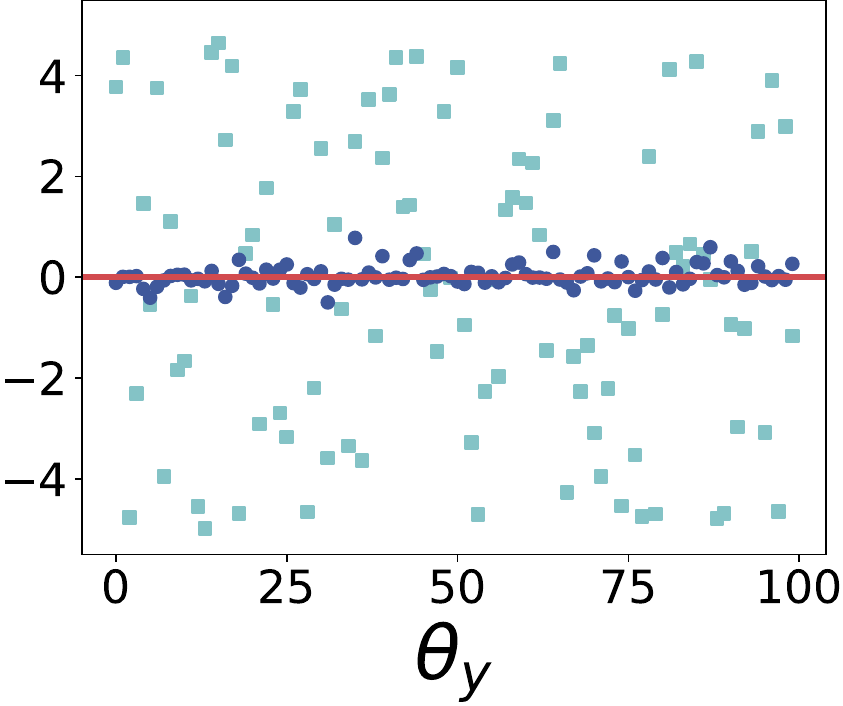}
        \includegraphics[width=0.32\linewidth]{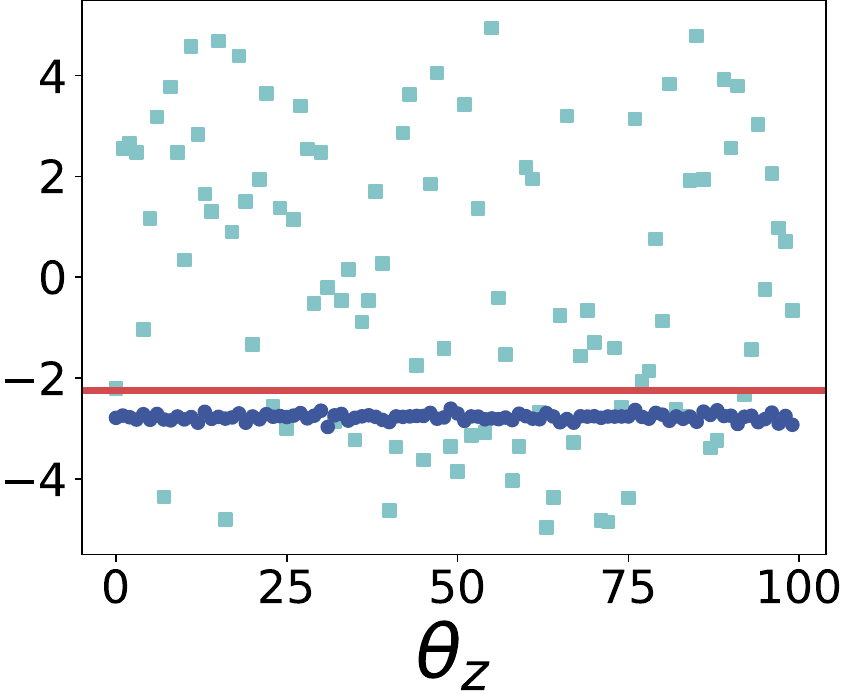}
    \end{subfigure}
    
    \vspace{3mm}
    
    \begin{subfigure}{1\linewidth}
        \centering
        \includegraphics[width=0.32\linewidth]{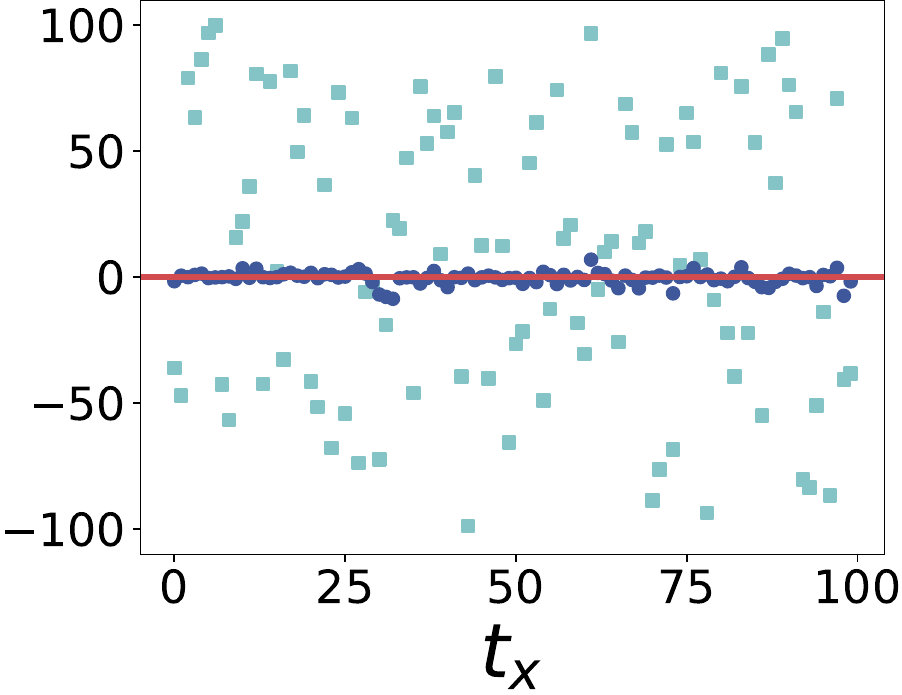}
        \includegraphics[width=0.32\linewidth]{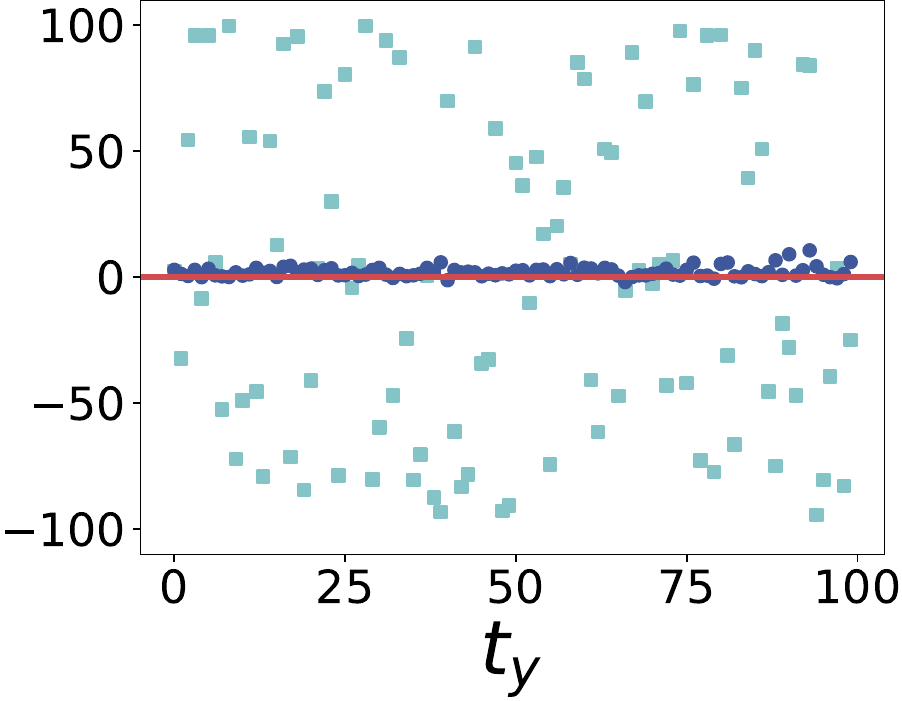}
        \includegraphics[width=0.32\linewidth]{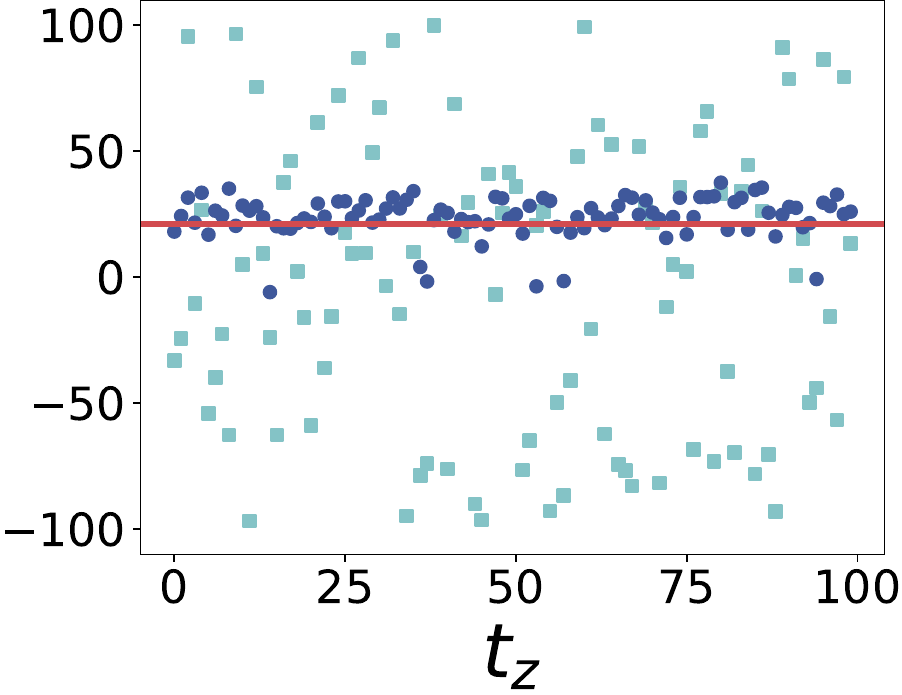}
    \vspace{-.2cm}
        \caption{ }
    \end{subfigure}

\caption{Initial parameters and the estimated results for each extrinsic in (a)~ORR and (b)~Boreas Dataset. The X-axis represents the index of the trial and the Y-axis is the value of the parameters in $[\theta] $ and $[cm]$ respectively.
\textcolor[RGB]{132,195,198}{Light blue} 
square indicates the initial value and the optimized result is in \textcolor[RGB]{63,88,155}{dark blue} point. The {\color{red}red} line shows the GT.
}
\label{fig:scatter}
\end{figure}
\subsection{Quantitative Results}

With the aforementioned settings, we apply all methods to the two datasets.
To mitigate effects caused by initial guesses for the optimization, we randomly generate initial parameters within the range of $[-5^\circ, 5^\circ]$ for rotation and $[-1~m, 1~m]$ for translation parameters.
Totally 100 initial sets are generated to evaluate the robustness of the optimization when the calibration parameters drift significantly from the ground truth.
We report the quantitative results of the average and standard deviation in Tab.~\ref{tab:orr_calibration} and \ref{tab:boreas_calibration} for the ORR and Boreas datasets, respectively.

\begin{figure}[t]
    \centering
\setlength{\belowcaptionskip}{-0.1cm}

        \begin{subfigure}{1\linewidth}
        \centering
        \setlength{\belowcaptionskip}{-0.1cm}
        \includegraphics[width=0.35\linewidth]{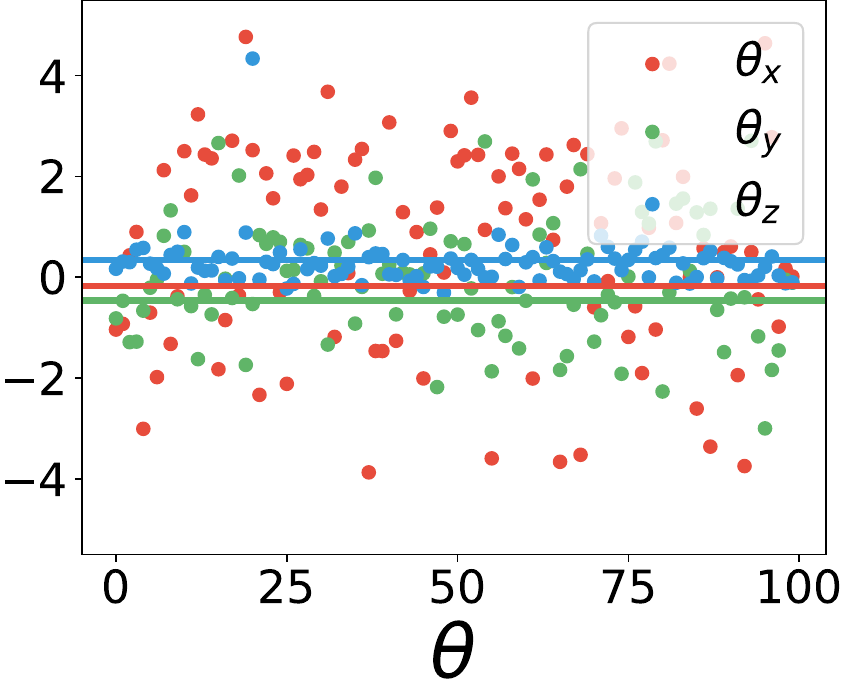}
        \hspace{3mm}
        \includegraphics[width=0.37\linewidth]{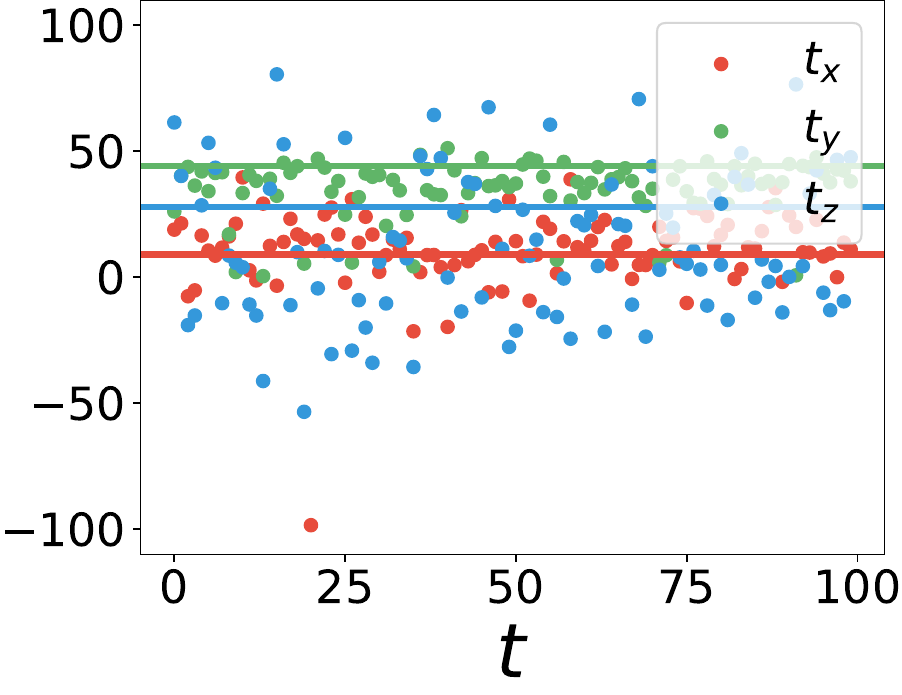}
    \vspace{-.4cm}
        \caption{ }
    \end{subfigure}

    \begin{subfigure}{1\linewidth}
        \centering
        \setlength{\belowcaptionskip}{-0.1cm}
        \includegraphics[width=0.35\linewidth]{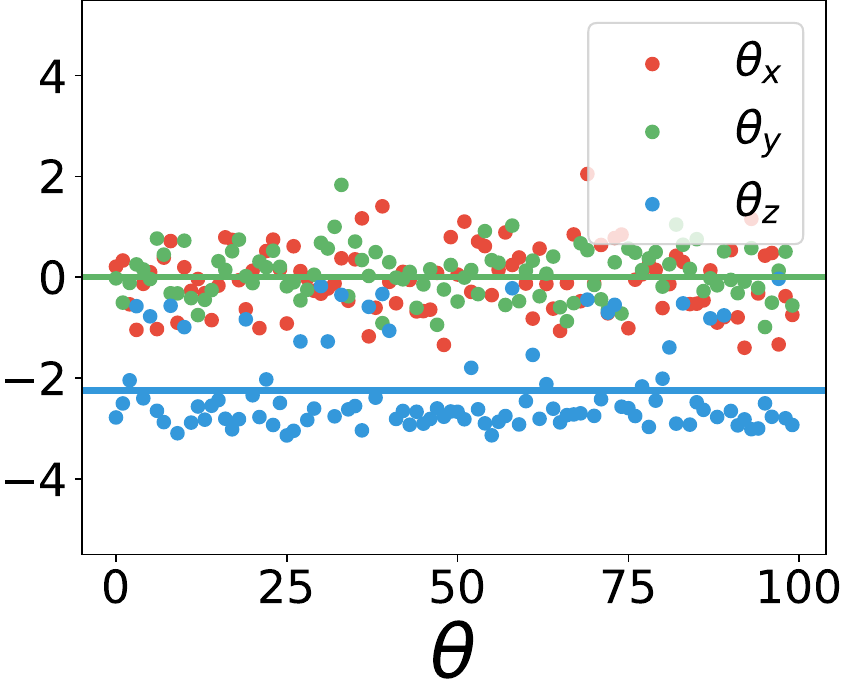}
        \hspace{3mm}
        \includegraphics[width=0.37\linewidth]{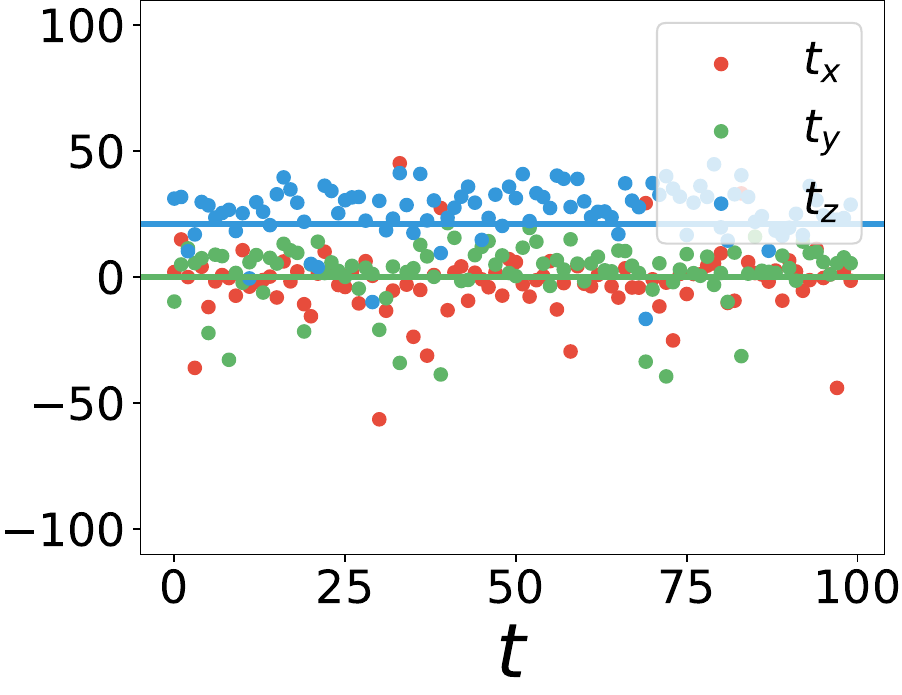}
    \vspace{-.4cm}
        \caption{ }
    \end{subfigure}
\caption{Estimated extrinsic parameters by LiRaCo with LiDAR point cloud data transformed by random noises for (a)~ORR and (b)~Boreas datasets. 
}
\label{fig:initial_noise}
\end{figure}
In both tables, it can be found that our proposed LiRaCo mostly outperforms all baseline methods in terms of accuracy and lower standard deviation, while also being capable of estimating all parameters.
\begin{figure*}[h!]
\setlength{\belowcaptionskip}{-0.1cm}
    \centering
    \begin{subfigure}{0.32\linewidth}
        \centering
        \includegraphics[width=1\linewidth]{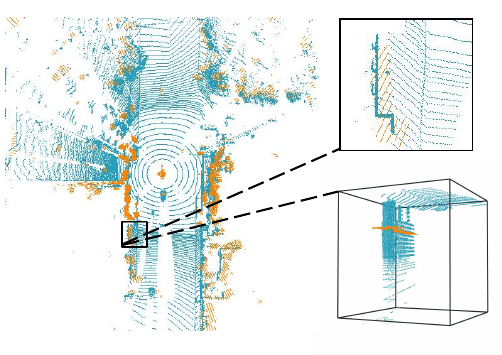}
                \vspace{-.7cm}
        \caption{}
    \end{subfigure}
    \centering
    \begin{subfigure}{0.32\linewidth}
        \centering
        \includegraphics[width=1\linewidth]{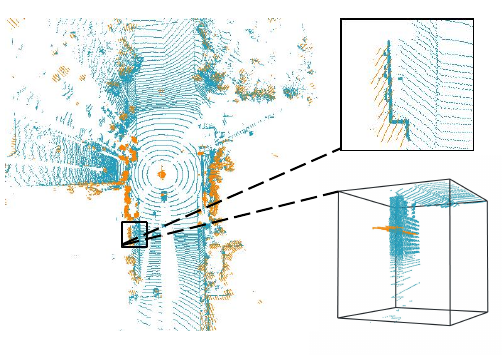}
                \vspace{-.7cm}
        \caption{}
    \end{subfigure}
    \begin{subfigure}{0.32\linewidth}
        \centering
        \includegraphics[width=1\linewidth]{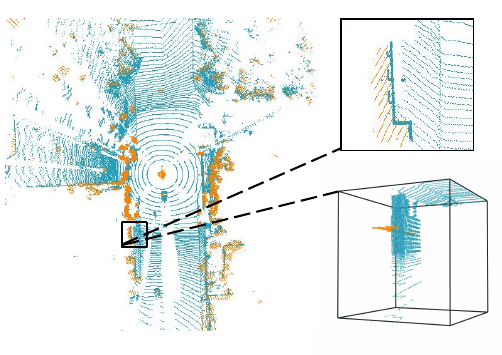}
                \vspace{-.7cm}
        \caption{}
    \end{subfigure}
 \caption{
 The top view and zoomed-in side view of the overlap between green LiDAR and brown Radar points using extrinsic parameters: (a) initial, (b) GT and (c) estimation by LiRaCo.
 }
\label{fig:vis_result}
\end{figure*}
Notably, LiRaCo demonstrates stable estimation ability even for the challenging height-related parameters: $\theta_x, \theta_y$ and $t_z$, despite the low vertical resolution of the data. This validates the practicality and effectiveness of our Radar 3D occupancy grid design.
Regarding the differences between the two datasets, LiRaCo exhibits higher accuracy on the height-related parameters $\theta_x$, $\theta_y$, and $t_z$ in the Boreas dataset. We attribute this improved performance to the utilization of the 128-beam LiDAR which brings richer height information in this dataset.

\begin{figure}[t]
\centering
\setlength{\belowcaptionskip}{-0.1cm}
 \begin{subfigure}{1\linewidth}
        \centering
        \includegraphics[width=0.45\linewidth]
        {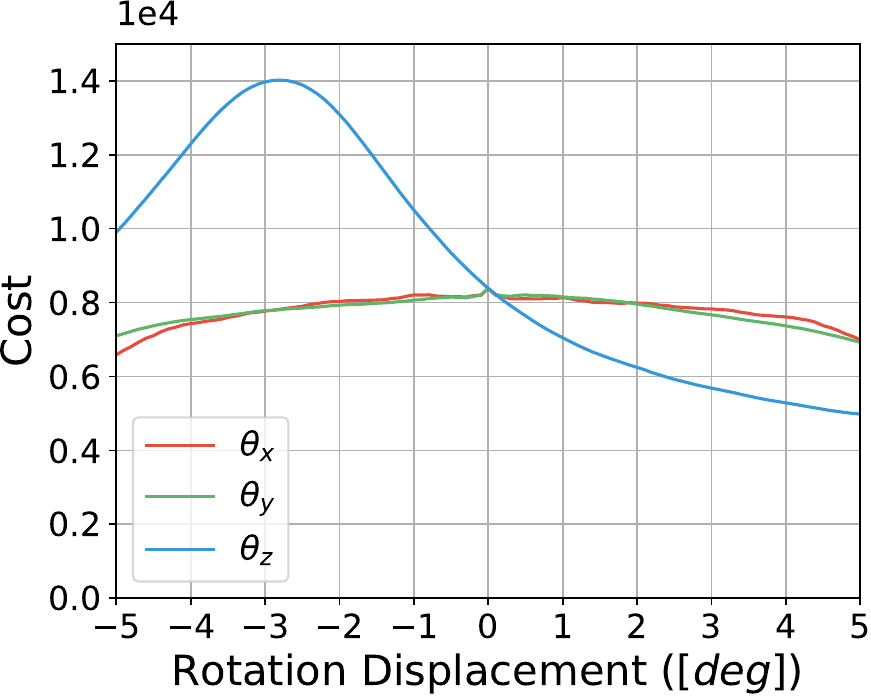}
        \includegraphics[width=0.45\linewidth]{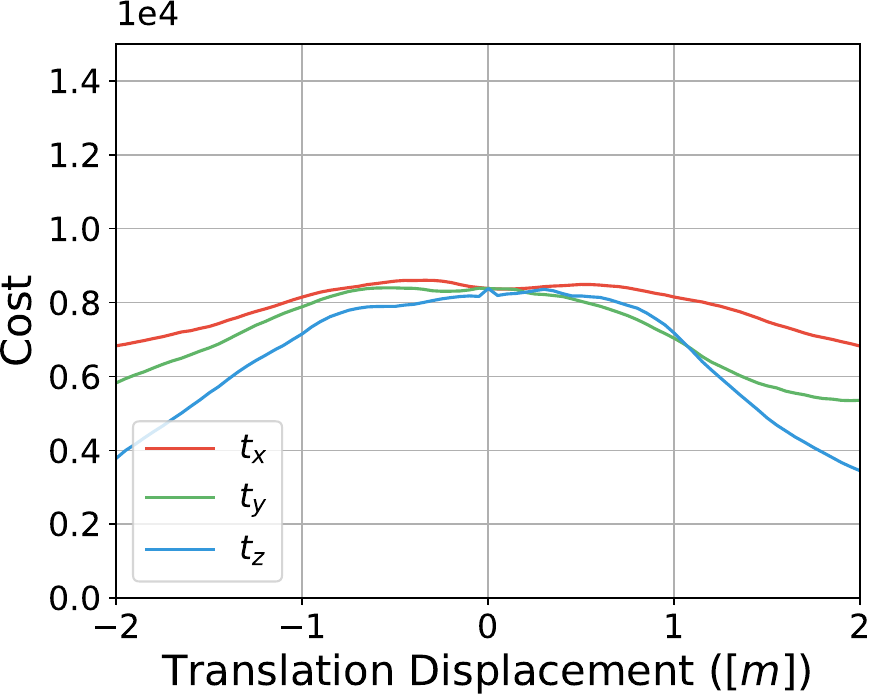}
        \caption{ }
        \label{fig:cost_fig_a}
\end{subfigure}
\vspace{-0.2cm}
\begin{subfigure}{1\linewidth}
        \centering
\includegraphics[width=0.45\linewidth]{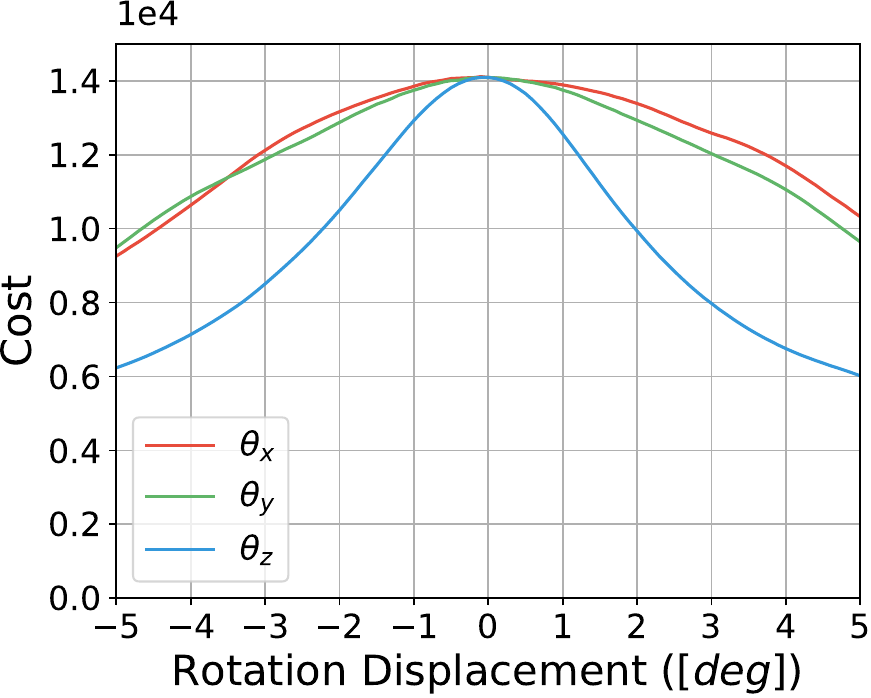}
\includegraphics[width=0.45\linewidth]{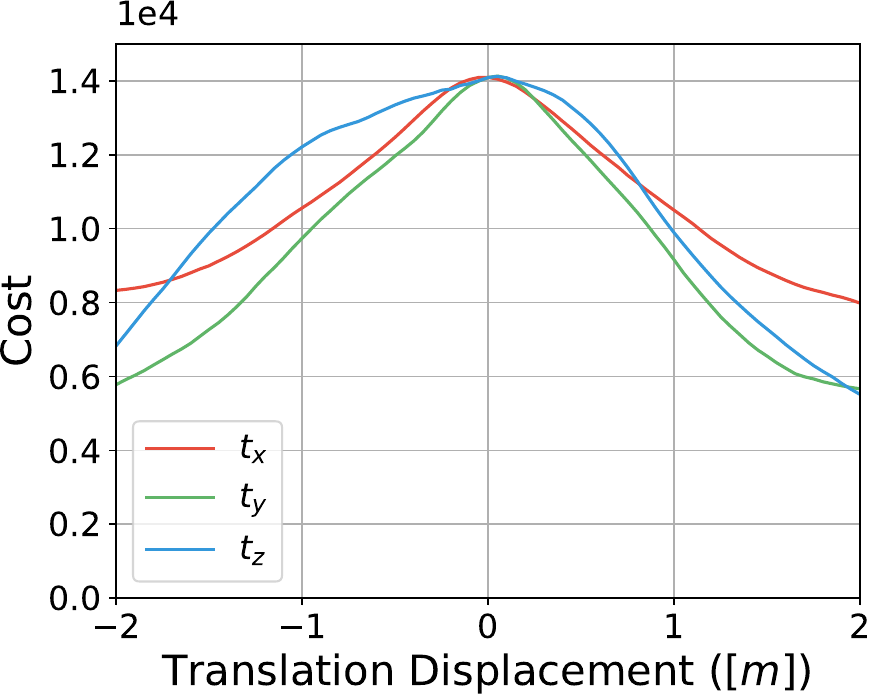}
\caption{}
    \end{subfigure}
\caption{The curves of the proposed cost function under varying rotation~($\theta_x,\theta_y,\theta_z$) and translation ~($t_x,t_y,t_z$) displacements in Boreas dataset. The X-axis represents the displacements of extrinsic parameters while the Y-axis indicates the corresponding cost. (a): cost curves for displacements based on the initial guess before calibration; (b): curves for displacement from the calibrated extrinsic parameters  estimated by LiRaCo.} 
\label{fig:cost_fig}
\end{figure}
To visually illustrate the robustness of LiRaCo to the initial guess, we plot the distribution of the initial value as light blue squares and the corresponding estimated results as dark blue scatters in Fig.~\ref{fig:scatter}.  The GT is marked with a red line. It can be observed that LiRaCo generally achieves stable estimations, particularly as the number of laser beams increases. Besides, it shows higher deviation for $\theta_x$, $\theta_y$, and $t_z$, indicating the difficulty in the estimation of the height-related parameters.

To further investigate the robustness to spatial variations of different sensor sets, we apply random transformation noise to the original LiDAR point cloud and report the calibrated results in Fig.~\ref{fig:initial_noise}.
The generation of random noise  follows the same process used for initial guesses.
The experiments also revealed that random noise poses greater challenges to convergence compared to random initial guesses for the optimization. This leads us to an enhanced optimization strategy to ensure the effectiveness and robustness of our approach.
For each set of random noise, we perform three optimization attempts: one starting from an initial value of 0 and two with random initial guesses to avoid local optima. The final extrinsic parameters are determined based on the optimization attempt with the lowest termination cost.
The results demonstrate that our method can handle various extrinsic parameter offsets, showing its potential and practicality for various real-world scenarios.

\begin{figure}[t]
\centering
        \centering
        \includegraphics[width=0.45\linewidth]{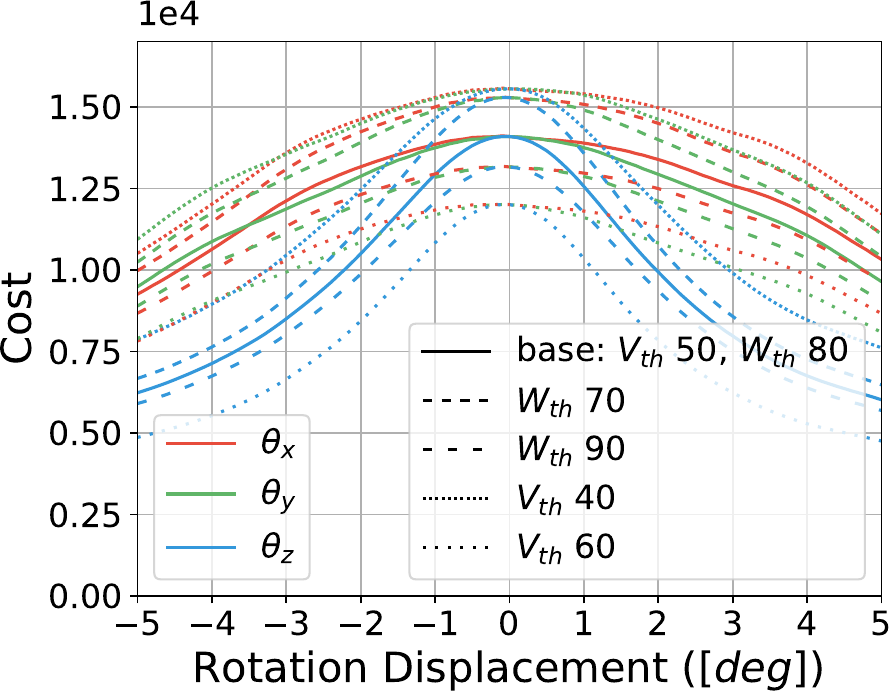}
        \includegraphics[width=0.45\linewidth]{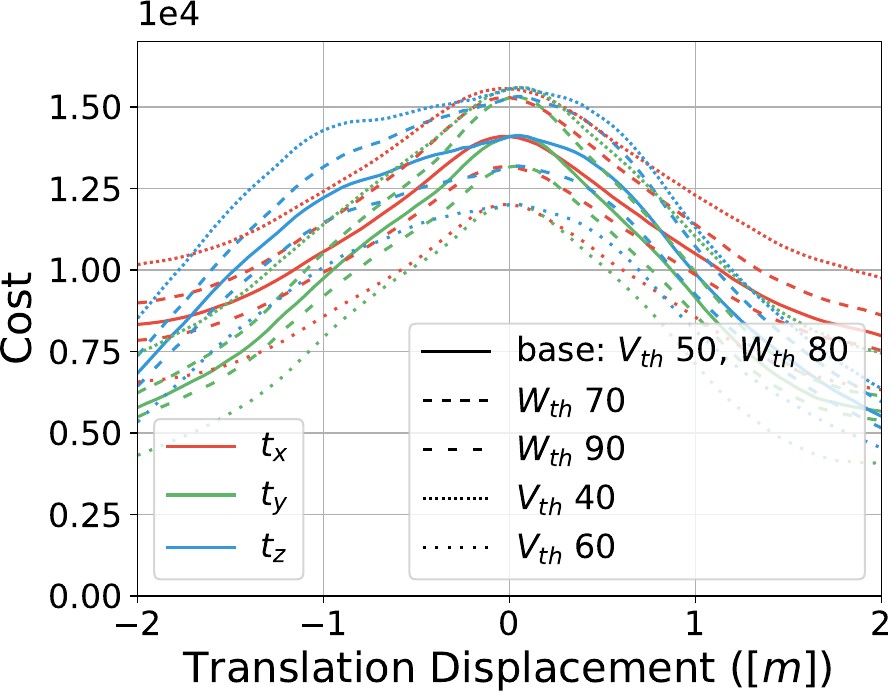}

\caption{The cost curves under different intensity thresholds $V_{th}$ and $W_{th}$ for Radar data. } 
\label{fig:cost_fig_intensity}
\end{figure}
\subsection{Qualitative Results}
We visualize the overlap with different extrinsic calibration parameters for qualitative comparison in Fig.~\ref{fig:vis_result}. 
A sample pair of LiDAR point clouds and the Radar points are projected onto the BEV images with notable areas zooming in for details. Prior to calibration, a significant offset between LiDAR and Radar can be observed, particularly at the corner of the wall and along the side of the road. With the calibration parameters by LiRaCo, the results demonstrate improved alignment between the two modalities from both the BEV and side view.

\subsection{Convergence Analysis}

Furthermore, to quantitatively validate the convexity of the cost function, we plot the cost curve with displacements from the initial and calibrated values of Boreas dataset in Fig.~\ref{fig:cost_fig}. 
The displacement ranges from $-5^{\circ}$ to $5^{\circ}$ with an interval of $0.1^{\circ}$  for rotation and  from $-2m$ to $2m$ with an interval of $0.05m$ for translations. It can be found that both curves are almost convex, especially the curve corresponding to the displacements from the calibrated values. This indicates that the estimated parameters are near the optimal values.

Additionally, we investigate the impact of different intensity thresholds $V_{th}$ for valid occupied grids and $W_{th}$ for weight enhanced grids, as introduced in \S~\ref{chap:3.3}.
The cost curves at different intensity thresholds are shown in Fig.~\ref{fig:cost_fig_intensity}.
The results indicate that the cost function maintains a stable convexity across a wide range of intensity thresholds, demonstrating the low sensitivity of our method to Radar intensity.
This suggests that these values exhibit strong transferability when applied to other similar real-world Radar sensor configurations.

\section{CONCLUSIONS}
In this paper, we proposed LiRaCo, a targetless LiDAR and 2D scanning Radar calibration method for estimating extrinsic parameters to enhance the robustness of sensor fusion. To establish dense spatial correspondences between the two modalities, we expand 2D Radar scans into 3D occupancy grids and align them with LiDAR points in the cylindrical coordinate system.
Consequently, we formulated the cost function that considers the overlap, vertical sparsity of LiDAR and intensity of Radar. The extrinsic parameters are derived by optimizing this cost function. The effectiveness and superiority of LiRaCo are validated by the comparisons with baseline methods and extensive ablation studies on real-scanned  ORR and Boreas datasets.

\addtolength{\textheight}{-12cm}   %

\bibliographystyle{IEEEtran}  %
\bibliography{IEEEabrv,ref}

\begin{thebibliography}{10}
\providecommand{\url}[1]{#1}
\csname url@rmstyle\endcsname
\providecommand{\newblock}{\relax}
\providecommand{\bibinfo}[2]{#2}
\providecommand\BIBentrySTDinterwordspacing{\spaceskip=0pt\relax}
\providecommand\BIBentryALTinterwordstretchfactor{4}
\providecommand\BIBentryALTinterwordspacing{\spaceskip=\fontdimen2\font plus
\BIBentryALTinterwordstretchfactor\fontdimen3\font minus
  \fontdimen4\font\relax}
\providecommand\BIBforeignlanguage[2]{{%
\expandafter\ifx\csname l@#1\endcsname\relax
\typeout{** WARNING: IEEEtran.bst: No hyphenation pattern has been}%
\typeout{** loaded for the language `#1'. Using the pattern for}%
\typeout{** the default language instead.}%
\else
\language=\csname l@#1\endcsname
\fi
#2}}

\bibitem{radar_different_representation}
S.~Yao, R.~Guan, Z.~Peng, C.~Xu, Y.~Shi, Y.~Yue, E.~G. Lim, H.~Seo, K.~L. Man,
  X.~Zhu, \emph{et~al.}, ``Radar perception in autonomous driving: Exploring
  different data representations,'' \emph{arXiv preprint arXiv:2312.04861},
  2023.

\bibitem{MVDnet}
K.~Qian, S.~Zhu, X.~Zhang, and L.~E. Li, ``Robust multimodal vehicle detection
  in foggy weather using complementary lidar and radar signals,'' in \emph{2021
  IEEE/CVF Conference on Computer Vision and Pattern Recognition (CVPR)}, 2021,
  pp. 444--453.

\bibitem{ST-mvdNet}
Y.-J. Li, M.~O’Toole, and K.~Kitani, ``St-mvdnet++: Improve vehicle detection
  with lidar-radar geometrical augmentation via self-training,'' in
  \emph{ICASSP 2023 - 2023 IEEE International Conference on Acoustics, Speech
  and Signal Processing (ICASSP)}, 2023, pp. 1--5.

\bibitem{RaLiBEV}
Y.~Yang, J.~Liu, T.~Huang, Q.-L. Han, G.~Ma, and B.~Zhu, ``Ralibev: Radar and
  lidar bev fusion learning for anchor box free object detection system,''
  \emph{arXiv preprint arXiv:2211.06108}, 2022.

\bibitem{Bi-LRFusion}
Y.~Wang, J.~Deng, Y.~Li, J.~Hu, C.~Liu, Y.~Zhang, J.~Ji, W.~Ouyang, and
  Y.~Zhang, ``Bi-lrfusion: Bi-directional lidar-radar fusion for 3d dynamic
  object detection,'' in \emph{Proceedings of the IEEE/CVF Conference on
  Computer Vision and Pattern Recognition (CVPR)}, June 2023, pp.
  13\,394--13\,403.

\bibitem{6DoF1}
J.~Per{\v{s}}i{\'c}, I.~Markovi{\'c}, and I.~Petrovi{\'c}, ``Extrinsic 6dof
  calibration of 3d lidar and radar,'' in \emph{2017 European Conference on
  Mobile Robots (ECMR)}.\hskip 1em plus 0.5em minus 0.4em\relax IEEE, 2017, pp.
  1--6.

\bibitem{tool1}
J.~Domhof, J.~F. Kooij, and D.~M. Gavrila, ``An extrinsic calibration tool for
  radar, camera and lidar,'' in \emph{2019 International Conference on Robotics
  and Automation (ICRA)}.\hskip 1em plus 0.5em minus 0.4em\relax IEEE, 2019,
  pp. 8107--8113.

\bibitem{extrinsic_temporal}
C.-L. Lee, Y.-H. Hsueh, C.-C. Wang, and W.-C. Lin, ``Extrinsic and temporal
  calibration of automotive radar and 3d lidar,'' in \emph{2020 IEEE/RSJ
  International Conference on Intelligent Robots and Systems (IROS)}.\hskip 1em
  plus 0.5em minus 0.4em\relax IEEE, 2020, pp. 9976--9983.

\bibitem{polarnet}
Y.~Zhang, Z.~Zhou, P.~David, X.~Yue, Z.~Xi, B.~Gong, and H.~Foroosh,
  ``Polarnet: An improved grid representation for online lidar point clouds
  semantic segmentation,'' 2020.

\bibitem{nie2023partner}
M.~Nie, Y.~Xue, C.~Wang, C.~Ye, H.~Xu, X.~Zhu, Q.~Huang, M.~B. Mi, X.~Wang, and
  L.~Zhang, ``Partner: Level up the polar representation for lidar 3d object
  detection,'' in \emph{Proceedings of the IEEE/CVF International Conference on
  Computer Vision}, 2023, pp. 3801--3813.

\bibitem{trust-region-methods}
A.~R. Conn, N.~I. Gould, and P.~L. Toint, \emph{Trust region methods}.\hskip
  1em plus 0.5em minus 0.4em\relax SIAM, 2000.

\bibitem{ST-MVDNet++}
Y.-J. Li, M.~O’Toole, and K.~Kitani, ``St-mvdnet++: Improve vehicle detection
  with lidar-radar geometrical augmentation via self-training,'' in
  \emph{ICASSP 2023 - 2023 IEEE International Conference on Acoustics, Speech
  and Signal Processing (ICASSP)}, 2023, pp. 1--5.

\bibitem{Radar_and_Lidar_Deep_Fusion}
Y.~Jin, Y.~Kuang, M.~Hoffmann, C.~Schüßler, A.~Deligiannis, J.-C.
  Fuentes-Michel, and M.~Vossiek, ``Radar and lidar deep fusion: Providing
  doppler contexts to time-of-flight lidar,'' \emph{IEEE Sensors Journal},
  vol.~23, no.~20, pp. 25\,587--25\,600, 2023.

\bibitem{orr_dataset}
D.~Barnes, M.~Gadd, P.~Murcutt, P.~Newman, and I.~Posner, ``The oxford radar
  robotcar dataset: A radar extension to the oxford robotcar dataset,'' in
  \emph{2020 IEEE International Conference on Robotics and Automation
  (ICRA)}.\hskip 1em plus 0.5em minus 0.4em\relax IEEE, 2020, pp. 6433--6438.

\bibitem{radiate_dataset}
M.~Sheeny, E.~De~Pellegrin, S.~Mukherjee, A.~Ahrabian, S.~Wang, and A.~Wallace,
  ``Radiate: A radar dataset for automotive perception in bad weather,'' in
  \emph{2021 IEEE International Conference on Robotics and Automation
  (ICRA)}.\hskip 1em plus 0.5em minus 0.4em\relax IEEE, 2021, pp. 1--7.

\bibitem{static_auto_calibration_3pairs}
S.~Agrawal, S.~Bhanderi, K.~Doycheva, and G.~Elger, ``Static multi-target-based
  auto-calibration of rgb cameras, 3d radar, and 3d lidar sensors,'' \emph{IEEE
  Sensors Journal}, 2023.

\bibitem{Extrinsics_RA_LI_learning}
P.~Jiang and S.~Saripalli, ``Improving extrinsics between radar and lidar using
  learning,'' \emph{arXiv preprint arXiv:2305.10594}, 2023.

\bibitem{lee2023extrinsic}
C.-L. Lee, C.-Y. Hou, C.-C. Wang, and W.-C. Lin, ``Extrinsic and temporal
  calibration of automotive radar and 3d lidar in factory and on-road
  calibration settings,'' \emph{IEEE Open Journal of Intelligent Transportation
  Systems}, 2023.

\bibitem{ILCC}
W.~Wang, K.~Sakurada, and N.~Kawaguchi, ``Reflectance intensity assisted
  automatic and accurate extrinsic calibration of 3d lidar and panoramic camera
  using a printed chessboard,'' \emph{Remote Sensing}, vol.~9, no.~8, 2017.

\bibitem{XU2021103776}
X.~Xu, L.~Zhang, J.~Yang, C.~Liu, Y.~Xiong, M.~Luo, Z.~Tan, and B.~Liu,
  ``Lidar–camera calibration method based on ranging statistical
  characteristics and improved ransac algorithm,'' \emph{Robotics and
  Autonomous Systems}, vol. 141, p. 103776, 2021.

\bibitem{10341781}
L.~F.~T. Fu, N.~Chebrolu, and M.~Fallon, ``Extrinsic calibration of camera to
  lidar using a differentiable checkerboard model,'' in \emph{2023 IEEE/RSJ
  International Conference on Intelligent Robots and Systems (IROS)}, 2023, pp.
  1825--1831.

\bibitem{6DoF2}
J.~Per{\v{s}}i{\'c}, I.~Markovi{\'c}, and I.~Petrovi{\'c}, ``Extrinsic 6dof
  calibration of a radar--lidar--camera system enhanced by radar cross section
  estimates evaluation,'' \emph{Robotics and Autonomous Systems}, vol. 114, pp.
  217--230, 2019.

\bibitem{tool2}
J.~Domhof, J.~F.~P. Kooij, and D.~M. Gavrila, ``A joint extrinsic calibration
  tool for radar, camera and lidar,'' \emph{IEEE Transactions on Intelligent
  Vehicles}, vol.~6, no.~3, pp. 571--582, 2021.

\bibitem{2D_radar-camera}
D.~Kim and S.~Kim, ``Extrinsic parameter calibration of 2d radar-camera using
  point matching and generative optimization,'' in \emph{2019 19th
  International Conference on Control, Automation and Systems (ICCAS)}.\hskip
  1em plus 0.5em minus 0.4em\relax IEEE, 2019, pp. 99--103.

\bibitem{3DRadar2ThermalCalib}
J.~Zhang, S.~Zhang, G.~Peng, H.~Zhang, and D.~Wang, ``3dradar2thermalcalib:
  Accurate extrinsic calibration between a 3d mmwave radar and a thermal camera
  using a spherical-trihedral,'' in \emph{2022 IEEE 25th International
  Conference on Intelligent Transportation Systems (ITSC)}, 2022, pp.
  2744--2749.

\bibitem{mutual_info}
G.~Pandey, J.~R. McBride, S.~Savarese, and R.~M. Eustice, ``Automatic extrinsic
  calibration of vision and lidar by maximizing mutual information,''
  \emph{Journal of Field Robotics}, vol.~32, no.~5, pp. 696--722, 2015.

\bibitem{wang2020soic}
W.~Wang, S.~Nobuhara, R.~Nakamura, and K.~Sakurada, ``Soic: Semantic online
  initialization and calibration for lidar and camera,'' \emph{arXiv preprint
  arXiv:2003.04260}, 2020.

\bibitem{sem_ins_cali}
C.~Sun, Z.~Wei, W.~Huang, Q.~Liu, and B.~Wang, ``Automatic targetless
  calibration for lidar and camera based on instance segmentation,'' \emph{IEEE
  Robotics and Automation Letters}, vol.~8, no.~2, pp. 981--988, 2023.

\bibitem{10356133}
Z.~Gong, R.~He, K.~Gao, and G.~Cai, ``Scene-aware online calibration of lidar
  and cameras for driving systems,'' \emph{IEEE Transactions on Instrumentation
  and Measurement}, pp. 1--1, 2023.

\bibitem{opencalib}
G.~Yan, Z.~Liu, C.~Wang, C.~Shi, P.~Wei, X.~Cai, T.~Ma, Z.~Liu, Z.~Zhong,
  Y.~Liu, \emph{et~al.}, ``Opencalib: A multi-sensor calibration toolbox for
  autonomous driving,'' \emph{Software Impacts}, vol.~14, p. 100393, 2022.

\bibitem{heng2020automatic}
L.~Heng, ``Automatic targetless extrinsic calibration of multiple 3d lidars and
  radars,'' in \emph{2020 IEEE/RSJ International Conference on Intelligent
  Robots and Systems (IROS)}.\hskip 1em plus 0.5em minus 0.4em\relax IEEE,
  2020, pp. 10\,669--10\,675.

\bibitem{iKalibr}
S.~Chen, X.~Li, S.~Li, Y.~Zhou, and X.~Yang, ``ikalibr: Unified targetless
  spatiotemporal calibration for resilient integrated inertial systems,''
  \emph{IEEE Transactions on Robotics}, pp. 1--20, 2025.

\bibitem{jung2024imaging}
S.~Jung, H.~Jang, M.~Jung, A.~Kim, and M.-H. Jeon, ``Imaging radar and lidar
  image translation for 3-dof extrinsic calibration,'' \emph{Intelligent
  Service Robotics}, vol.~17, no.~2, pp. 167--179, 2024.

\bibitem{boreas_dataset}
K.~Burnett, D.~J. Yoon, Y.~Wu, A.~Z. Li, H.~Zhang, S.~Lu, J.~Qian, W.-K. Tseng,
  A.~Lambert, K.~Y. Leung, \emph{et~al.}, ``Boreas: A multi-season autonomous
  driving dataset,'' \emph{The International Journal of Robotics Research},
  vol.~42, no. 1-2, pp. 33--42, 2023.

\bibitem{opencv_library}
G.~Bradski, ``{The OpenCV Library},'' \emph{Dr. Dobb's Journal of Software
  Tools}, 2000.

\bibitem{icp}
P.~Besl and N.~D. McKay, ``A method for registration of 3-d shapes,''
  \emph{IEEE Transactions on Pattern Analysis and Machine Intelligence},
  vol.~14, no.~2, pp. 239--256, 1992.

\end{thebibliography}

\end{document}